\documentclass[11pt, a4paper, onecolumn, copyright, goog]{google}
\usepackage{subcaption}
\usepackage{comment}
\usepackage{multirow}
\usepackage{hyperref}
\hypersetup{
    colorlinks=true,
    linkcolor=blue,
    urlcolor=blue,
}

\usepackage[utf8]{inputenc}
\usepackage{xcolor} 

\uselogo{} 

\usepackage{kantlipsum, lipsum}
\usepackage{dm-colors}
\usepackage{amsmath}
\usepackage{pstricks, pst-node}
\usepackage{verbatim}
\usepackage{multirow}
\usepackage{scalerel}
\usepackage{longtable}
\usepackage{array}  
\usepackage{booktabs}
\usepackage{enumitem}
\usepackage{xspace}
\usepackage{bm}
\usepackage{bbm}
\usepackage{mathtools}
\usepackage{soul}
\usepackage{epsfig}
\usepackage{graphicx}
\usepackage{tcolorbox}
\usepackage{subcaption}
\usepackage{amssymb}
\usepackage{colortbl}
\usepackage{csquotes}
\usepackage{etex}
\usepackage{setspace}
\usepackage{svg}
\usepackage{colortbl}
\usepackage{tabularx,ragged2e}
\usepackage{placeins}
\usepackage{afterpage}
\usepackage[symbol]{footmisc}
\usepackage[bibstyle=nature,citestyle=numeric-comp,%
            natbib=true,backend=biber,maxbibnames=99,%
            giveninits=false,sorting=none]{biblatex}
\usepackage{nameref}
\usepackage{varioref}
\usepackage[noabbrev,capitalize]{cleveref}

\usepackage{etoc}
\usepackage{lineno}
\graphicspath{{figures/}}

\title{Towards Expert-level Medical AI\\for Real-time Video Consultations}

\author[$\ast$,1]{Mahvish Nagda}
\author[$\ast$,1]{Jihyeon Lee}
\author[1]{Matthew Thompson}
\author[2]{Chunjong Park}
\author[2]{Tim Strother}
\author[2]{Valentin Liévin}
\author[1]{\\Roma Ruparel}
\author[1]{Akshay Goel}
\author[1]{Teya Bergamaschi}
\author[1]{Suhana Bedi}
\author[2]{Meet Shah}
\author[2]{Pavel Dubov}
\author[2]{Liviu Panait}
\author[2]{\\Toshiyuki Fukuzawa}
\author[2]{Sam Schmidgall}
\author[1]{Craig Schiff}
\author[1]{Joseph Xu}
\author[2]{Aliya Rysbek}
\author[2]{Yana Lunts}
\author[2]{Jan Freyberg}
\author[1]{\\Rebecca Hemengway}
\author[1]{Sunny Virmani}
\author[2]{David Racz}
\author[2]{Carey Radebaugh}
\author[2]{Joëlle Barral}
\author[1]{Kavi Goel}
\author[1]{\\Dale R. Webster}
\author[1]{Katherine Chou}
\author[1]{Avinatan Hassidim}
\author[1]{Yossi Matias}
\author[1]{James Manyika}
\author[2]{\\Gregory Wayne}
\author[2]{Tao Tu}
\author[1]{Yun Liu}
\author[1]{Ethan Goh}
\author[1]{Christina Chen}
\author[2]{Ryutaro Tanno}
\author[1]{\\Po-Hsuan Cameron Chen}
\author[$\dagger$,1]{Mike Schaekermann}
\author[$\dagger$,1]{Anil Palepu}

\affil[*]{Equal contributions}
\affil[$\dagger$]{Equal leadership}

\affil[1]{Google Research, }
\affil[2]{Google DeepMind}

\correspondingauthor{$\{$apalepu$,\,$mikeshake$\}$@google.com}
\begin{document}

\begin{refsection}

\begin{abstract}

Audio-visual interaction is the standard for patient-physician consultations, enabling natural communication and effective assessment of illness through non-verbal cues. While text-based AI has shown promise, it discards essential perceptual dimensions and limits patients who cannot articulate symptoms in writing. Early efforts to extend medical AI to audio-visual interaction have demonstrated feasibility, but not reached clinician-level performance. Here, we provide the first demonstration of expert-level AI in real-time clinical video consultations using AMIE (Articulate Medical Intelligence Explorer) in a video configuration. AMIE (Video) is a Gemini-based multi-agent system integrating low-latency dialogue, clinical reasoning, and real-time audio-visual perception. To guide development, we established a taxonomy and automated evaluations for clinical audio-visual cues in telehealth settings. In a randomized Objective Structured Clinical Examination (OSCE) study with 30 primary care physicians (PCPs), 15 patient actors and 100 clinical scenarios, we compared AMIE (Video), its text-only counterpart AMIE (Text), and PCPs consulting via video. Clinical evaluators rated AMIE (Video) on par or better than PCPs in history-taking, diagnosis, management, and physical observation and examination. Patient actors preferred AMIE's approach to assessing and explaining conditions, while PCPs were preferred for rapport and partnership building. In modality ablation, patient actors preferred AMIE (Video)'s interface over text chat for communicative effectiveness, convenience, and feeling understood. Limitations remain in fine anatomical precision, subtle affective nuances, and high-frequency movements. While further research is needed before real-world translation, these results mark an important milestone toward AI systems capable of augmenting care across the sensory complexity of clinical practice.

\end{abstract}

\maketitle

\section{Introduction}
\label{sec:introduction}

The information exchanged in a clinical encounter extends far beyond what words alone can convey.
Throughout a consultation, skilled physicians continuously integrate non-verbal cues, including vocal tone, facial expression, posture, and the display of physical signs such as dyspnea or tremor \cite{hall1995nonverbal}, alongside the spoken history.
These visual and auditory dimensions are foundational to clinical reasoning \cite{soriano2025bates}, shaping the trust, empathy, and rapport that underpin the therapeutic relationship \cite{henry2012association}, and are intrinsically linked to clinical outcomes including symptom resolution and patient adherence \cite{stewart1995effective}.
Evaluating physical appearance, behavior, and posture is a core component of clinical examination \cite{wright2022assessment}, and these cues remain important to patient satisfaction during telehealth encounters \cite{kruis2024patient}, where many physical examination maneuvers can be successfully adapted to video-based settings \cite{benziger2021telehealth, tong2026translatability}.

\begin{figure}[ht!]
  \centering
  \includegraphics[width=\textwidth]{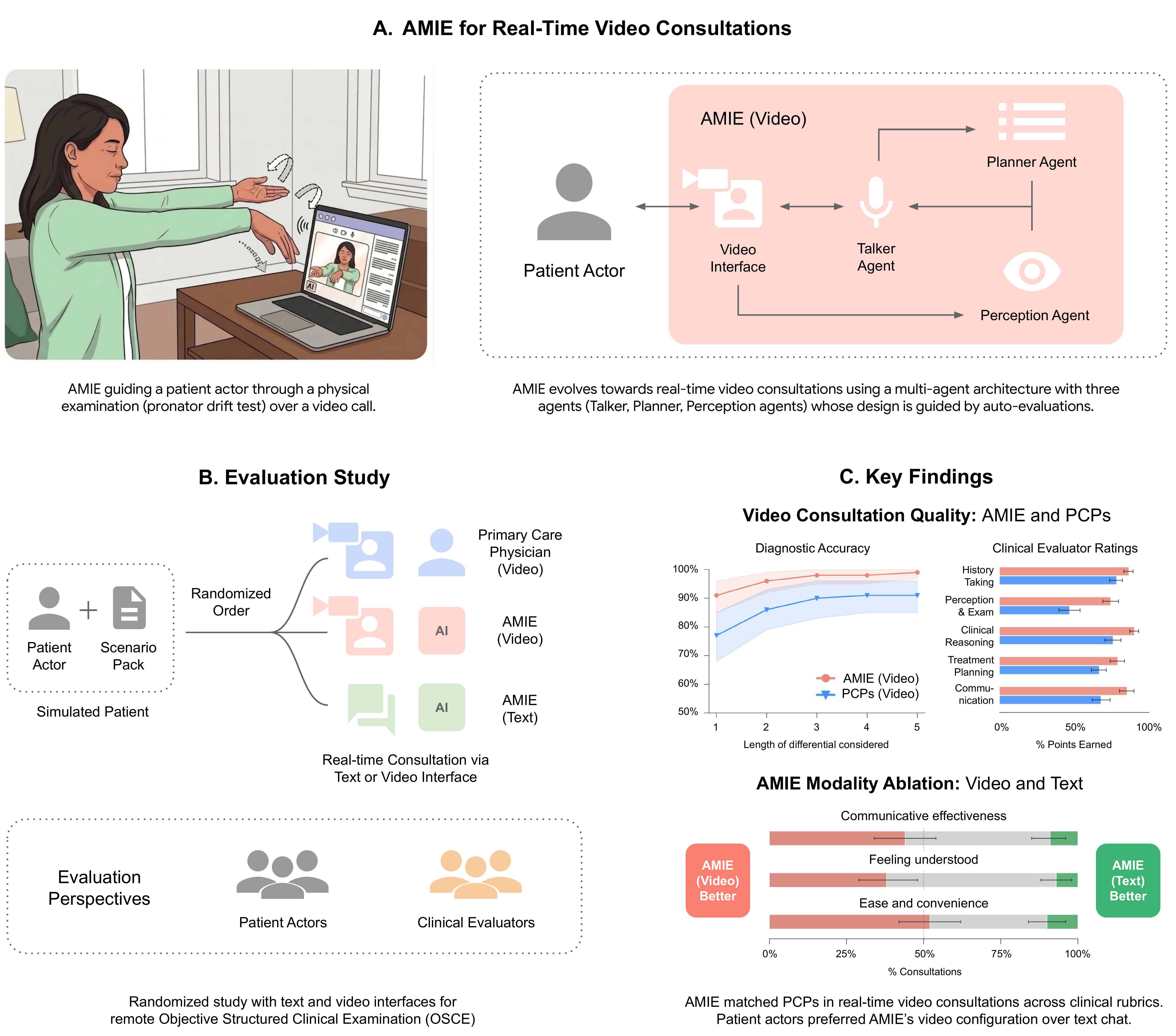}
  \vspace{0.1cm}
  \caption{\textbf{Overview of Contributions.} This work provides the first demonstration of expert-level AI performance in real-time clinical video consultations, enabled by three key contributions: \textbf{(A)} AMIE (Video), a Gemini-based asynchronous multi-agent architecture comprising a Talker, Planner, and Perception agent, guided by an automated evaluation framework derived from a taxonomy of clinical audio-visual competencies; \textbf{(B)} A multi-arm randomized video-based OSCE comparing AMIE (Video), AMIE (Text), and PCPs (Video) across 100 clinical scenarios enacted by professional patient actors, and evaluated by independent clinical evaluators using standard OSCE criteria and case-specific rubrics; \textbf{(C)} Clinical evaluators rated AMIE (Video) on par or better than PCPs across core clinical competencies, and patient actors preferred the video interface over text chat.}
  \label{fig:contributions_overview}
\end{figure}

Video-based consultations are the dominant modality for remote clinical care, making them an ecologically valid setting for evaluating clinical AI.
Yet expert-level capabilities of AI for clinical reasoning and dialogue have been demonstrated exclusively in instant messaging settings, matching or exceeding human performance in text-based conversations \cite{mcduff2025towards, lievin2026towards, tu2025towards} and in settings where additional information is shared through images \cite{saab2026advancing}.
Early prospective evidence suggests conversational safety and clinical utility of text-based patient-facing AI in real-world clinical studies \cite{korom2025ai,zeltzer2025comparison,brodeur2026prospectiveclinicalfeasibilitystudy}.
Even with the addition of images, these interfaces cannot fully capture the dynamic audio-visual cues of a live clinical encounter, and patients still need to distill complex physical and behavioral observations into discrete descriptions, a process that introduces biases \cite{sackett1979bias} and risks discarding diagnostic information.
Text-only interfaces also present barriers to equitable access, potentially excluding patients with lower digital or health literacy who may struggle to articulate symptoms in writing \cite{arias2023digital, moore2025patient}.

Early efforts to extend medical AI to audio-visual interaction have demonstrated feasibility \cite{shah2026towards}, but not reached clinician-level performance, and key challenges remain.
An effective audio-visual clinical agent must balance low-latency dialogue with thorough clinical reasoning and real-time audio-visual perception while maintaining robust performance across diverse patient encounters.
Fully characterizing these systems demands larger-scale evaluation, including modality ablations with text-based counterparts, and granular analyses of key clinical audio-visual capabilities.

Here, we provide the first demonstration of expert-level AI performance in real-time clinical video consultations, presenting the Articulate Medical Intelligence Explorer (AMIE) in a video configuration.
This advance is enabled by three key contributions (\cref{fig:contributions_overview}):
\begin{itemize}
    \item \textbf{AMIE for Real-Time Video Consultations:} We introduce AMIE in its video configuration, AMIE (Video), a novel Gemini-based asynchronous multi-agent architecture that overcomes the latency, perception, and robustness challenges inherent in real-time, video-based clinical interactions. To characterize system capabilities and failure modes, we derived a taxonomy of clinical audio-visual competencies feasible in telehealth settings from the medical literature, and developed a corresponding automated evaluation suite comprising targeted single-turn assessments and multi-turn simulated consultations. Through this framework, we demonstrate how the individual components of AMIE (Video) contribute to overall system performance.
    \item \textbf{Multi-arm Video OSCE Study:} In a randomized Objective Structured Clinical Examination (OSCE) study, we compared AMIE (Video), its text-only counterpart AMIE (Text), and 10 board-certified primary care physicians (PCPs) conducting video consultations across 100 clinical scenarios with 15 professional patient actors, evaluated by an independent panel of 20 clinical evaluator PCPs. Consultation quality is evaluated from both the patient actor and clinical evaluator perspective, including standard OSCE assessment criteria as well as case-specific rubrics designed for each clinical scenario.
    \item \textbf{Key Findings:} Clinical evaluators rated AMIE (Video) on par or better than PCPs in history-taking, diagnosis, management, and physical observation and examination. Patient actors preferred AMIE's approach to assessing and explaining conditions, while PCPs were preferred for rapport and partnership building. In modality ablation, patient actors preferred AMIE's video interface over text chat for communicative effectiveness, convenience, and feeling understood.
\end{itemize}
\section{Methods}
\label{sec:methods}

We developed AMIE (Video) to conduct clinical consultations through a real-time video interface as is routine practice in many telehealth settings. The system coordinates several specialized agents (\cref{fig:agent_design}, \cref{sec:amie}): the user-facing \textit{Talker Agent} (\cref{sec:methods:talker}) generates rapid patient-facing responses, informed by the background \textit{Planner Agent} (\cref{sec:methods:planner_agent}) managing clinical goals, and the \textit{Perception Agent} (\cref{sec:methods:perception_agent}) processing continuous audio and video streams. To guide development, we established an automated evaluation framework (\cref{sec:autoeval}) covering various audio-visual capabilities (\cref{sec:methods:evaluation_capabilities}) via both single-turn assessments (\cref{sec:methods:agent_evaluation}) and simulated multi-turn conversations (\cref{sec:methods:multiturn_evaluation}). Finally, we conducted a randomized multi-arm OSCE study (\cref{sec:osce_study}) comparing text and video configurations of AMIE with a group of primary care physicians (PCPs) (\cref{sec:methods:pcps}) across 100 simulated patients (\cref{sec:methods:patient_actors}). Consultations were assessed on general and case-specific rubrics (\cref{sec:case_specific_rubrics}) by a panel of clinical evaluators (\cref{sec:methods:evaluators}). 

\subsection{System Architecture}
\label{sec:amie}

\begin{figure}[ht!]
  \centering
  \includegraphics[width=\textwidth]{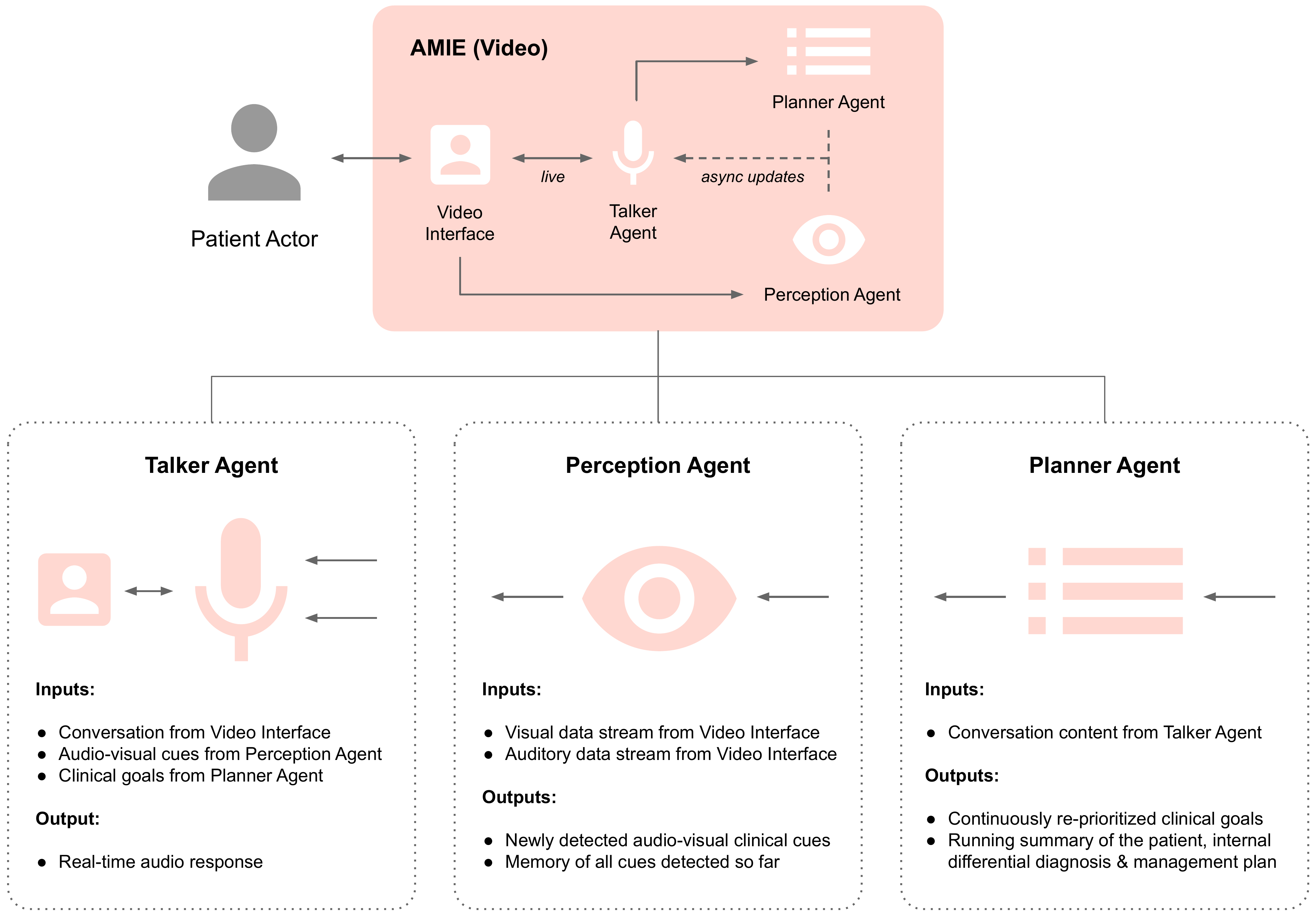}
  \vspace{0.1cm}
  \caption{\textbf{AMIE for Real-Time Video Consultations.} AMIE in its video configuration, AMIE (Video), balances low-latency dialogue with thorough clinical reasoning and real-time audio-visual perception via a three-agent architecture interfacing with patient actors through a video interface. The \textbf{Talker Agent} serves as the conversational interface of the system and aims to deliver low latency responses to the patient. It pulls the latest updated state from the other two agents to guide its responses. The \textbf{Perception Agent} is tasked to identify clinically relevant audio-visual cues within the context of the given conversation. It also records a cumulative list of observations, which helps account for the limited prior video context seen by the system at any given moment. The \textbf{Planner Agent} manages the clinical goals of the system, adding, removing, and re-prioritizing goals as needed as the conversation unfolds. It also synthesizes information received into concise summaries of patient information and the system's internal diagnostic and management reasoning.}
  \label{fig:agent_design}
\end{figure}

AMIE (Video) (\cref{fig:agent_design}) is an experimental multi-agent conversational system designed to conduct real-time video-based clinical dialogue with the goal to generate clinically appropriate differential diagnoses and management plans. The system tested in this work is built upon Gemini 3 Flash~\cite{gemini3flash}  and Gemini 3.1 Pro~\cite{gemini3p1pro}. The system relies on real-time orchestration of three specialized agents collaborating asynchronously to deliver fluent responses with low latency informed by thorough clinical reasoning and live audio-visual perception.

\subsubsection{Talker Agent}
\label{sec:methods:talker}

The Talker Agent is tasked with responding fluently to user utterances throughout the conversation, requiring low-latency response generation. Building upon the Project Astra infrastructure and Gemini's native multimodal understanding, the agent processes multimodal conversations to generate real-time audio responses. When the user speaks, the Talker listens. When the Talker detects a sufficient period of silence following a user utterance, it triggers a model call to generate a response to the user. Even while speaking, the Talker continues to listen for user input, and transitions to silent listening mode upon detecting an interruption from the user. This approach affords natural conversational turn-taking between the user and the agent. To minimize latency, the Talker utilizes a minimal compute budget and processes only the most recent 5 seconds of video frames when formulating its immediate response.
Finally, the Talker serves as the primary orchestrator of the system, consolidating inputs from the other two agents. For response generation, it synthesizes the most recent states from the Perception Agent (observations of audio-visual clues captured so far) and the Planner Agent (current clinical goals and reasoning), alongside recent video frames and the full dialogue history.

\subsubsection{Planner Agent}
\label{sec:methods:planner_agent}

The Planner Agent manages the system's clinical goals. Its design was inspired by the \textit{Management Reasoning (Mx) Agent} from a prior iteration of AMIE \cite{lievin2026towards} and further expands upon the implementation described in \cite{shah2026towards}. The Planner maintains a persistent memory of the clinical state, including a summary of patient symptoms, a differential diagnosis, a management plan, and active conversational `milestones'. These milestones may specify information to collect (e.g., ``elicit the exact onset, character, and radiation of the chest pain''), physical actions for the patient to perform (e.g., ``instruct the patient to demonstrate active shoulder range of motion''), or critical guidance to provide (e.g., ``advise escalation to urgent care if shortness of breath worsens''). Upon invocation, the Planner executes a multi-step reasoning chain via structured decoding, pruning, re-prioritizing, or appending new goals while updating its understanding of the clinical case.
To maintain fluency, updates to the Planner's internal state take place asynchronously, decoupled from activity of the Talker Agent, which responds promptly based on the most recent Planner state. State updates are triggered whenever new user input is received to minimize information staleness, subject to a rate limit of one deep reasoning invocation per 10 seconds of dialogue to balance model load.
Several measures help to mitigate potential misalignment between the Planner's state and the Talker's context. First, the Planner focuses on high-level clinical goals which update only occasionally throughout the conversation. Second, the Talker is designed through prompting to handle potential discrepancies. Finally, the system continues to process incoming information while the Talker delivers audio output to the user.

\subsubsection{Perception Agent}
\label{sec:methods:perception_agent}

The Perception Agent is designed to perceive and interpret real-time audio and video streams, using Gemini 3.1 Pro in a dedicated inference call to carefully enumerate clinically relevant visual and auditory information. During conversational turn-taking, audio-visual cues, such as a patient's affect, speech cadence, or subtle physical signs, can occasionally go unaddressed by the Talker itself given the reasoning budget constraints necessitated by its priority on low-latency fluent dialogue. The Perception Agent addresses this limitation by reasoning over an extended video window, and maintaining a persistent memory of all clinical audio-visual cues detected over time. For example, if a patient exhibits a dry cough at the beginning of an interaction, the Perception Agent will log this observation and  ensure it remains accessible to the system for subsequent clinical reasoning.
The Perception Agent implements the same asynchronous relationship with the Talker Agent as described for the Planner Agent above.

\subsection{Automated Evaluation}
\label{sec:autoeval}

To guide development, we constructed an automated evaluation framework designed to assess system performance across a spectrum of clinical and conversational contexts relevant to real-time video telehealth settings (\cref{sec:methods:evaluation_capabilities}). The framework comprises two core paradigms: (1) targeted single-turn evaluations (\cref{sec:methods:agent_evaluation}), which isolate specific audio-visual capabilities; (2) simulated multi-turn interactions (\cref{sec:methods:multiturn_evaluation}), which assess end-to-end conversational capability as a proxy for eventual performance in the evaluation study.

\subsubsection{Telehealth Evaluation Taxonomy}
\label{sec:methods:evaluation_capabilities}

We first derived a taxonomy of clinical audio-visual cues and physical examination maneuvers relevant to telehealth settings (\cref{appendix:av_cues}). Synthesized from the medical literature and physical examination protocols \cite{McDonoughUWPhysicalExam, caltrc2021telehealth, ansary2021virtual,dover2023macleod}, our taxonomy encompasses physical observations and guided examinations across major body systems and observation categories. We categorized the clinical audio-visual cues by their ``telehealth feasibility''---whether they can be reasonably observed in a virtual setting---and by acquisition method, distinguishing between passive observations and those involving guided or elicited maneuvers.

For the purpose of this study, we further classified clinical signs into ``actable'' presentations (realistically simulated by patient actors, such as motor or behavioral changes) and ``non-actable'' presentations (specific physical or behavioral abnormalities that cannot be enacted).

\subsubsection{Single-turn Evaluation}
\label{sec:methods:agent_evaluation}
Single-turn agent evaluations served as `unit tests' for specific audio-visual telehealth capabilities, including clinical perception, reasoning, and interaction. These scenarios were grounded in audio or video clips (sourced from publicly available materials and study-specific enactments) and designed to elicit specific internal reasoning or dialogue responses from the agent at precise points in a patient conversation. For instance, an evaluation scenario testing visual perception may depict a patient raising their right hand and asking, ``what am I showing you?'' The target response requires accurate identification of both anatomical location and laterality. Each evaluation case was paired with example-specific clinical grading rubrics that describe the desired response and internal reasoning of the agent. During evaluation, an LLM-based auto-rater grades the agent's response with a quality score from 0 to 1 applying the rubric. 
Further details on the design of single-turn agent evaluations are described in \cref{appendix:agent_eval_design}.

\subsubsection{Multi-turn Evaluation}
\label{sec:methods:multiturn_evaluation}

While single-turn evaluation facilitates targeted benchmarking of perception, clinical encounters are inherently dynamic. For instance, while an agent should consistently elicit the anatomical location of a patient's abdominal pain, the precise turn at which this request occurs may vary across conversations. To evaluate these conversational-level competencies, we leveraged multi-turn evaluations, evolving techniques from prior work with text-based simulated AI patients \cite{tu2025towards,lievin2026towards}, to an \textit{audio} setting. In these simulations, visual physical examination cues were conveyed through injected verbal stage directions. For instance, if the agent instructs the patient to palpate their abdomen, the simulator might respond with the spoken audio confirmation, ``Sure,'' accompanied by the caption, ``[patient presses on their lower right abdomen while simultaneously wincing].'' 
This multi-turn evaluation was applied to the clinical scenarios used in \citet{shah2026towards}, and an LLM-based auto-rater graded the clinical dialogue using case-specific rubrics across core clinical dimensions (history-taking, clinical reasoning, communication and counseling, treatment planning, triage, and red-flag identification).

\begin{figure}[h!]
    \centering
    \includegraphics[width=\textwidth]{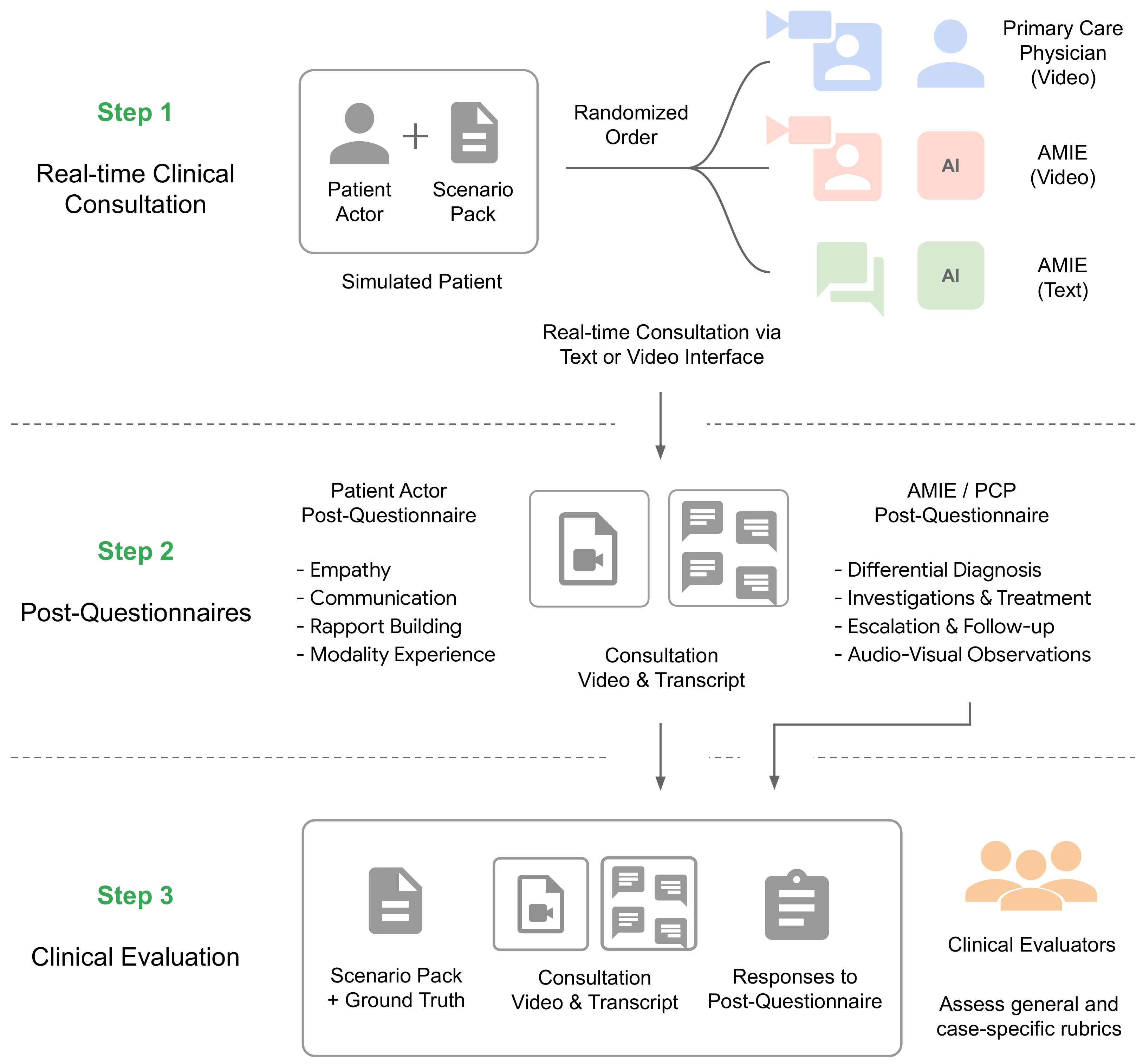}
    \caption{\textbf{Study Design.} A randomized Objective Structured Clinical Examination (OSCE) study compared AMIE (Video), its text-only counterpart AMIE (Text), and PCPs consulting via video, on the task performing a real-time consultation with simulated patients (professional patient actors enacting a given clinical scenario pack). After each consultation, AMIE and PCPs produced answers to a post-questionnaire documenting their clinical reasoning, and patient actors rated their experience on pre-defined rubrics.
    A separate panel of clinical evaluators assessed each consultation based on general and scenario-specific criteria.}
    \label{fig:study_design}
\end{figure}

\subsection{OSCE Study}
\label{sec:osce_study}

We conducted a randomized multi-arm evaluation study comparing AMIE with a group of US board-certified primary care physicians (PCPs) for the task of conducting real-time video-based telehealth consultations with patient actors, including one study arm testing AMIE in its text chat configuration (\cref{fig:study_design}). We adapted the virtual Objective Structured Clinical Examination (OSCE) framework introduced in \citet{tu2025towards}, designing a total of 120 scenario packs enriched with audio-visual cues and guided physical examination components. Complete study details, participant characteristics, and rubric descriptions are provided in \cref{appendix:osce_study_details}.

\subsubsection{Study Design and Clinical Scenarios}

A validation set of 20 clinical scenarios was used to conduct a study pilot with six arms comparing AMIE and PCPs in Video, Audio, and Text modalities each respectively, resulting in 120 completed consultations (6 across each of 20 scenarios).
The purpose of this study pilot was to perform an initial validation of the agent and interface, and to inform the final study design through a small-scale ablation isolating effects of text, audio, and audio-video modalities.
Details and results from the pilot are provided in \cref{appendix:pilot_run}.

The full evaluation study was conducted across a distinct set of 100 clinical scenarios across three arms: AMIE (Video), AMIE (Text), and PCP (Video). All arms were conducted through a teleconferencing interface resembling Google Meet affording interactions through text chat and real-time audio-video modalities. (\cref{figure:patient_actor_ui}).
In this study, AMIE did not feature a visual avatar.
For controlled comparison and to avoid bias from visual presence or appearance, PCPs were asked to turn their cameras off throughout the consultations (\cref{sec:study_design}). 

Scenario packs and simulated patients were prepared by an OSCE laboratory in North America and verified by US board-certified PCPs (\cref{sec:methods:scenario_packs}). Drawing from HHS resources for virtual care\footnote{https://telehealth.hhs.gov/providers/preparing-patients-for-telehealth/telehealth-physical-exam}, scenarios were designed to span five clinical domains relevant to telehealth settings: (1) cardiopulmonary, (2) abdominal, (3) HEENT (head, eyes, ear, nose, throat), (4) neurology and psychiatry, and (5) musculoskeletal. Each of these categories included a subset of non-disease cases with resolved complaints to assess clinical reasoning and behavior in those negative settings. Inpatient, pediatric, and dermatological cases, as well as those with privacy-sensitive physical examinations were excluded. Each pack defined ground-truth diagnoses, plausible alternatives, investigations, management plans, and a 20-point rubric used to score clinical capability in case-specific manner  (\cref{appendix:case_specific_rubric_example}).

Following each encounter, AMIE and PCPs produced answers to a structured post-questionnaire documenting differential diagnoses, observations, and management recommendations (\cref{box:post_questionnaire,sec:post_questionnaires}). For this post-encounter step, which operates under less stringent latency constraints than live dialogue, the AMIE system leveraged inference-time compute scaling via multi-draft synthesis \cite{lievin2026towards} to improve recommendation quality.

\subsubsection{Patient Actors and Primary Care Physicians}

Clinical scenarios were enacted by a cohort of 15 professional patient actors (7 female, 8 male; ages 20--75) trained to deliver standardized portrayals of clinical affect, physical cues, and guided physical examination maneuvers matched to biological sex (\cref{sec:methods:patient_actors}). Consultations in the PCP (Video) arm were performed by 10 US board-certified PCPs (mean 4.5 years post-residency clinical experience, \cref{sec:methods:pcps}). 

\begin{figure}[htp!]
    \vspace{-2em}
    \makebox[\textwidth][c]{\includegraphics[width=1.04\textwidth]{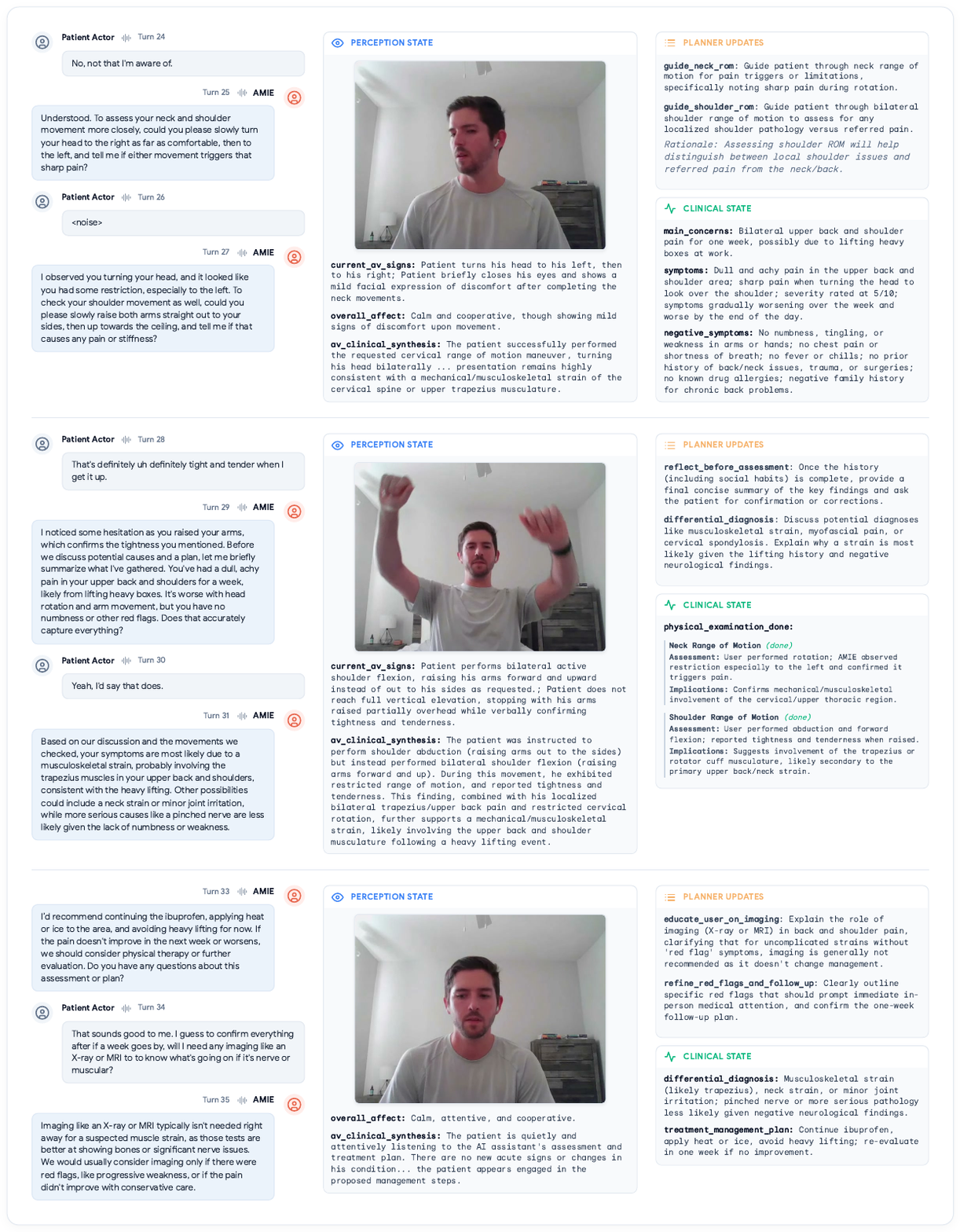}}
    \captionsetup{skip=0pt}
    \caption{\textbf{AMIE Interaction Example} showcasing activities of Talker, Perception and Planner agents.}
    \label{fig:illustrative_example}
\end{figure}

\subsubsection{Evaluation Rubrics}

Consultations were evaluated from multiple perspectives. Following each visit, patient actors evaluated conversational quality and communication convenience using established criteria from GMCPQ (General Medical Council Patient Questionnaire), PACES (Practical Assessment of Clinical Examination Skills), and PCCBP (Patient-Centered Communication Best Practice) adapted from \citet{tu2025towards} (\cref{sec:patient_actor_rubrics}). Additionally, they rated their experience with the given communication method (text chat or video interface, \cref{sec:patient_actor_rubrics}). 

A panel of 20 board-certified PCPs served as clinical evaluators (\cref{sec:methods:evaluators}) to assess consultation recordings, transcripts, and post-questionnaire responses for each scenario and study arm in a randomized order using a standard rating interface (\cref{figure:evaluator_ui}). 
Clinical evaluators graded encounters across both general and case-specific rubrics. General criteria, adapted from \citet{tu2025towards} and \citet{lievin2026towards}, assessed differential diagnosis (DDx) accuracy, diagnostic and management reasoning, and competencies in clinical dialogue (PACES and PCCBP), supplemented by telehealth-specific rubrics evaluating perceptual acuity and guided physical examination execution \cite{sartori2020telehealth} (\cref{sec:general_rubrics,appendix:av_telehealth_rubric}). Finally, each consultation was evaluated against a 20-point checklist rubric specific to each scenario spanning five domains: history taking, perception and examination, clinical reasoning, treatment planning, and communication. Details are provided in \cref{appendix:case_specific_rubric_example,sec:case_specific_rubrics}.

\subsubsection{Qualitative Data}
\label{sec:methods:qualitative}
Patient actors and clinical evaluators were encouraged to leave free-text comments in their assessments to provide specific rationale for their ratings. To further characterize consultation experience from the patient actor perspective, semi-structured interviews were conducted with 7 patient actors, probing at differences in interaction modalities and conversational experience. Qualitative data was analyzed using an reflexive thematic
analysis to inductively generate themes based on the transcripts and researcher notes. Details are provided in \cref{appendix:patient_actor_interviews}.

\subsubsection{Statistical Analysis}
\label{sec:methods:statistical_analysis}

For the 20-point case-specific rubrics, performance was aggregated as the mean score earned out of the possible points achievable across each of the five core rubric categories, as well as overall.
For ease of interpretation, Likert ratings from general assessment rubrics were standardized to a percentage scale for reporting (0\% representing the lowest ratings for all 100 cases, and 100\% representing the highest possible scores for all 100 cases).
Differences across study arms were evaluated using a two-stage procedure: an omnibus Friedman test was first conducted to assess for differences across all three arms, followed by two-sided Wilcoxon signed-rank tests and Benjamini-Hochberg False Discovery Rate (FDR) correction for comparisons between pairs of arms. 
Friedman and Wilcoxon signed-rank tests were conducted using scenario-level blocking.
Cases with N/A ratings on either arm of a comparison were excluded. All $p$ values reported in the manuscript refer to FDR adjusted p values from Wilcoxon signed rank tests.
Error bars correspond to 95\% confidence intervals produced via bootstrap procedure with 1000 resamples.

\section{Results}
\label{sec:results}

\begin{figure}[ht!]
    \centering
    \includegraphics[width=\textwidth]{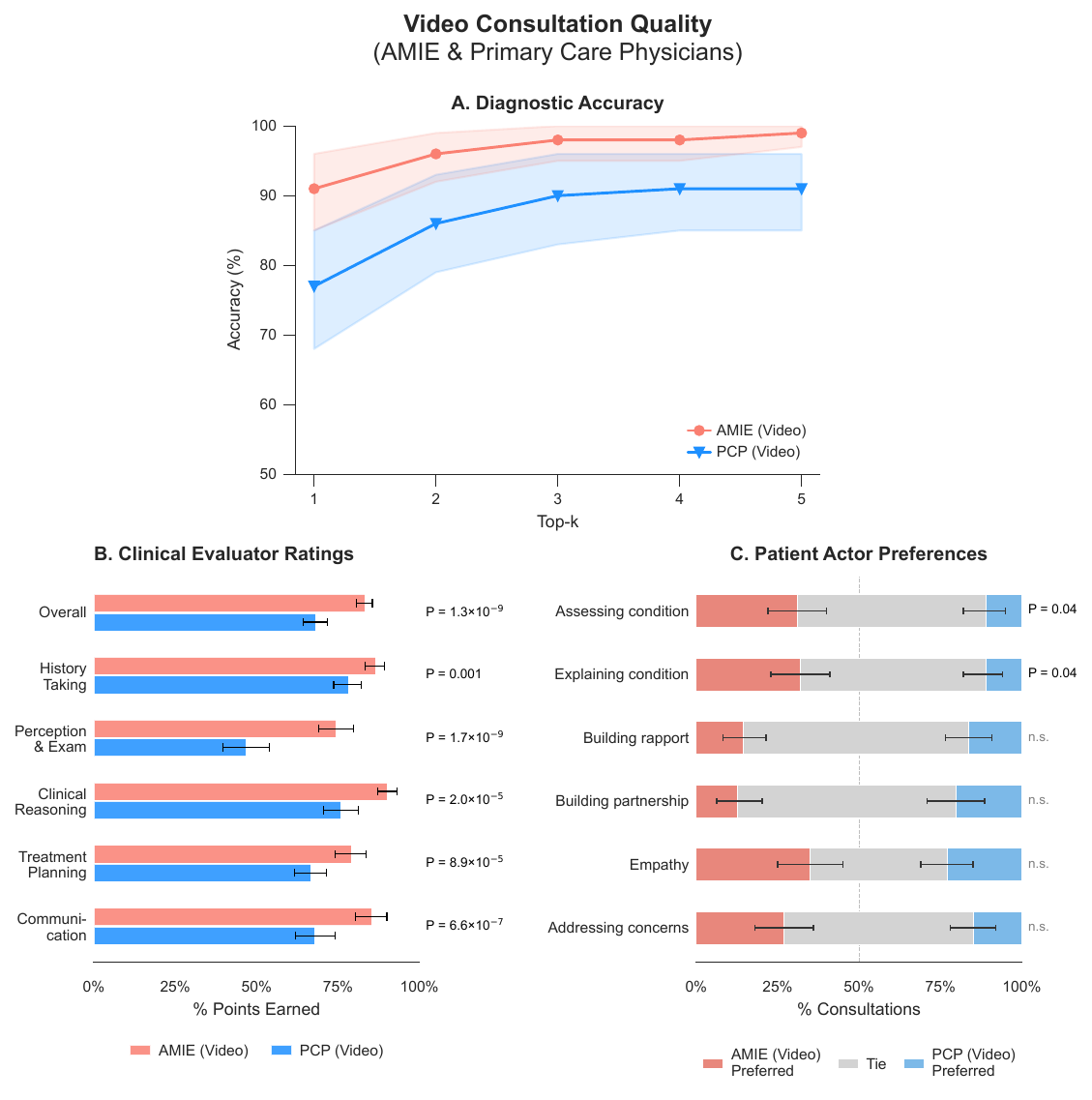}
    \caption{\textbf{Video Consultation Quality for AMIE and Primary Care Physicians.} Results are based on the OSCE evaluation study with patient actors and clinical evaluators ($N = 100$ clinical scenarios).
    \textbf{(A)} Top-$k$ cumulative diagnostic accuracy relative to each scenario's pre-specified reference diagnosis, where $k$ refers to the length of the differential diagnosis list considered.
    \textbf{(B)} Mean percentage of points earned on case-specific clinical rubrics across six consultation domains, as rated by clinical evaluators.
    \textbf{(C)} Patient actor preferences on six global criteria, comprising two items each selected from the GMCPQ (\textit{Assessing condition}, \textit{Explaining condition}), PCCBP (\textit{Building rapport}, \textit{Building partnership}), and PACES (\textit{Empathy}, \textit{Addressing concerns}) rubrics.
    Error bars indicate 95\% non-parametric bootstrap confidence intervals (1,000 resamples). $P$-values reflect false discovery rate (FDR) adjustments after Wilcoxon signed-rank tests (\textit{n.s.}\ denotes $P > 0.05$).}
    \label{fig:osce_main_results}
\end{figure}

\begin{figure}[ht!]
    \centering
    \includegraphics[width=\textwidth]{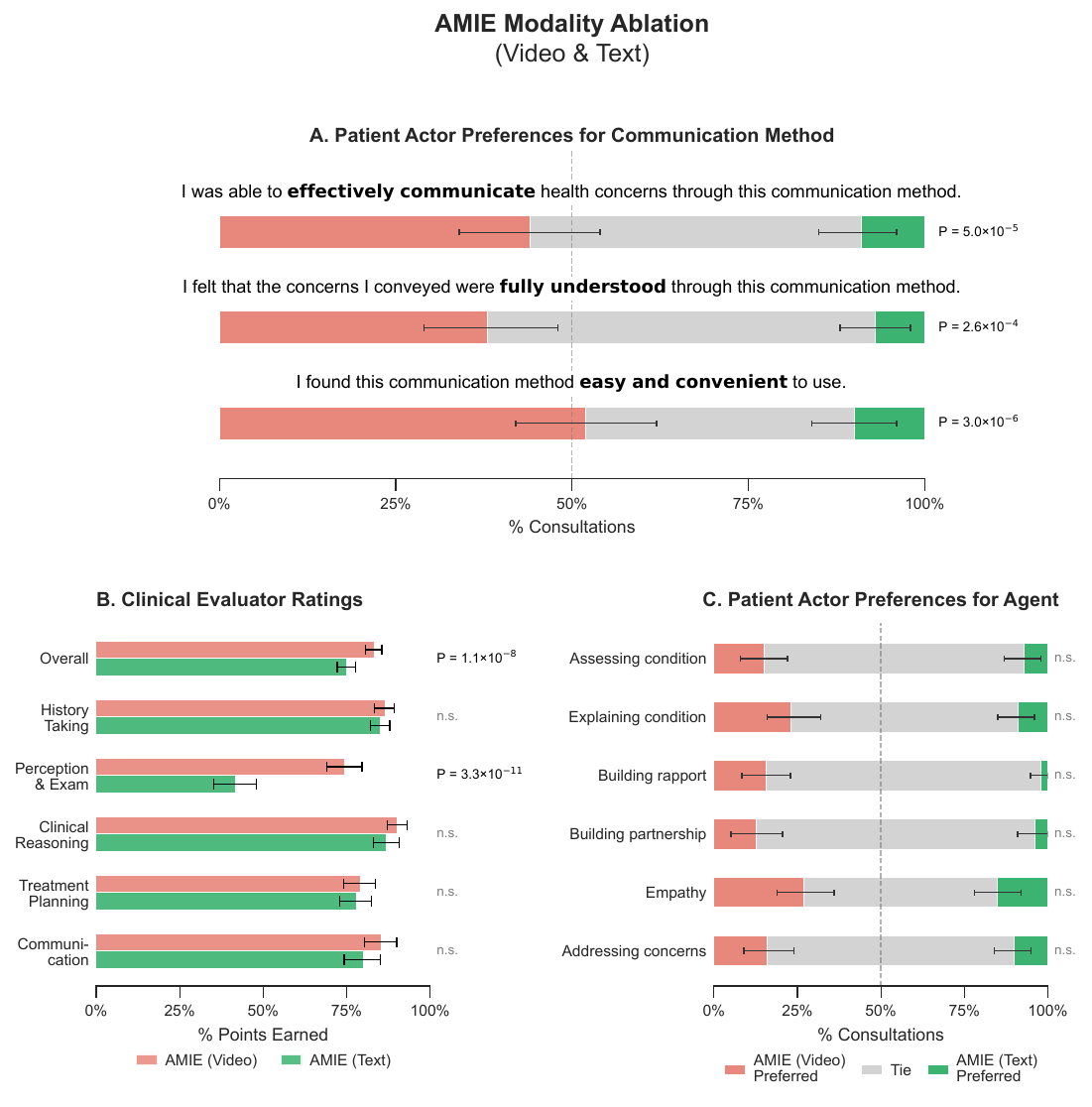}
    \caption{\textbf{AMIE Modality Ablation (Video \& Text).} Results are based on the OSCE evaluation study with patient actors and clinical evaluators ($N = 100$ clinical scenarios).
    \textbf{(A)} Patient actor ratings of the communication modality across three criteria.
    \textbf{(B)} Mean percentage of points earned on case-specific clinical rubrics across six consultation domains, as rated by clinical evaluators.
    \textbf{(C)} Patient actor preferences on six global criteria, comprising two items each selected from the GMCPQ (\textit{Assessing condition}, \textit{Explaining condition}), PCCBP (\textit{Building rapport}, \textit{Building partnership}), and PACES (\textit{Empathy}, 
    \textit{Addressing concerns}) rubrics. 
    Error bars indicate 95\% non-parametric bootstrap confidence intervals (1,000 resamples). $P$-values reflect false discovery rate (FDR) adjustments after Wilcoxon signed-rank tests (\textit{n.s.}\ denotes $P > 0.05$).}
    \label{fig:osce_modality_ablation}
\end{figure}

We describe key results from our comparison of video consultation quality between AMIE and PCPs (\cref{fig:osce_main_results}) and the modality ablation comparing AMIE between video and text chat interfaces  (\cref{fig:osce_modality_ablation}), as well as qualitative insights and results from automated evaluations below.
Excerpts from a representative AMIE (Video) consultation, including corresponding planner and perception agent states, are shown in \cref{fig:illustrative_example}. 
Full tabular results, scenario-level preferences, and pilot study results 
are provided in  \cref{appendix:tabular_results,appendix:full_patient_actor_ratings,%
appendix:full_expert_eval_ratings,appendix:pilot_run}.

\subsection{Video Consultation Quality: AMIE and PCPs}
\label{sec:results:amie_vs_pcp}

\textit{AMIE demonstrated expert-level performance in real-time 
video consultations:}
Clinical evaluators rated AMIE (Video) equivalent to or better than PCPs across all six case-specific rubric domains (\cref{fig:osce_main_results}B), with an overall score of 83\% compared with 68\% for PCPs ($p = 1.32 \times 10^{-9}$, \cref{tab:osce_case_rubric_results}).
Patient actors preferred AMIE (Video) for assessing and explaining conditions, while differences on empathy and addressing concerns were not statistically significant (\cref{fig:osce_main_results}C).

\textit{Diagnostic accuracy and clinical reasoning were higher than for PCPs:}
AMIE (Video) matched the correct reference diagnosis for each clinical scenario in its top item of the generated differential diagnosis in 91\% of cases, compared with 77\% for PCPs ($p = 0.039$). Diagnostic accuracy became more similar when considering the top three items in the differential for AMIE (Video) and PCPs respectively (98\% vs. 90\%, $p = 0.179$, \cref{fig:osce_main_results}A). On case-specific rubrics, AMIE (Video) scored 90\% compared with 76\% for PCPs on clinical reasoning ($p = 2.00 \times 10^{-5}$), and 79\% compared with 67\% on treatment planning ($p = 8.90 \times 10^{-5}$).

\textit{History-taking was rated equivalent to or better than PCPs:} 
Clinical evaluators rated AMIE (Video)'s history-taking favorably across established PACES criteria. On case-specific rubrics, AMIE (Video) scored 86\% compared with 78\% for PCPs ($p = 1.20 \times 10^{-3}$).  The mean conversation duration was comparable across both study arms (8.94 minutes for AMIE Video vs. 9.30 minutes for PCPs), suggesting a similar timeframe for information gathering.

\textit{AMIE leveraged video effectively for physical observation and guided examination:}
AMIE (Video) scored higher than PCPs at utilizing live video to augment information gathering (77\% vs. 51\%, $p = 1.37 \times 10^{-4}$) and guiding patients through physical examinations (72\% vs 39\%, $p = 2.46 \times 10^{-9}$).
These findings were reflected in case-specific perception and examination rubrics, where AMIE (Video) scored 74\% compared with 47\% for PCPs ($p = 1.73 \times 10^{-9}$).

\textit{Communication was rated favorably, though PCPs retained 
advantages in rapport:}
On established PACES and PCCBP scales, clinical evaluators rated empathy (82\% vs. 71\%, $p = 3.39 \times 10^{-3}$), fostering of relationships (80\% vs. 69\%, $p = 5.63 \times 10^{-3}$), and case-specific communication (85\% vs. 68\%, $p = 6.56 \times 10^{-7}$) for AMIE (Video) as equal to or better than PCPs.
Patient actors preferred AMIE (Video) for assessing and explaining conditions, while PCPs were preferred (though without a significant difference) for building rapport and partnership.

\subsection{AMIE Modality Ablation: Video and Text}
\label{sec:results:modality_ablation}

\textit{Patient actors preferred the video interface for communication:} 
Across all three communication method criteria (\cref{fig:osce_modality_ablation}A), patient actors rated AMIE (Video) significantly higher than AMIE (Text), including on effectiveness in communicating health concerns (89\% vs 79\%, $p = 4.98 \times 10^{-5}$), ease and convenience of use (88\% vs 71\%, $p = 2.98 \times 10^{-6}$), and feeling that their concerns were fully understood (90\% vs 81\%, $p = 2.62 \times 10^{-4}$).

\textit{Video conferred advantages in perception and examination:} 
Clinical evaluators rated AMIE (Video) higher than AMIE (Text) on overall case-specific rubrics (83\% vs 75\%, $p = 1.19 \times 10^{-8}$), with the largest gains in the perception and examination domain (\cref{fig:osce_modality_ablation}B). 
Diagnostic accuracy and management appropriateness were comparable across both modalities.

\textit{Agent-level preferences were largely similar across modalities:} 
Patient actor ratings of agent behavior on individual conversational criteria did not differ significantly between AMIE (Video) and AMIE (Text) (\cref{fig:osce_modality_ablation}C). While not significant, directionally AMIE (Video) scored higher than AMIE (Text) on all criteria, as reflected by the rating for ``Happy to see again'' (91\% vs 74\%, ($p = 0.038$).

\subsection{Qualitative Insights}
\label{sec:results:qualitative}

Semi-structured interviews with patient actors (\cref{appendix:patient_actor_interviews}, \cref{tab:qual-themes}) and free-text clinical evaluator comments revealed themes in how participants experienced AMIE (Video) relative to PCPs and across communication modalities.

\textit{Video enabled direct demonstration of physical symptoms:}
Patient actors generally preferred video for physical complaints, noting that it allowed them to show rather than describe symptoms.
As one actor explained, video enabled demonstrating specific movements versus \textit{``describing like I'm supinating my arm and it hurts.
You can just show them and then they can interpret that''} (P6).
Conversely, text was preferred for sensitive topics where auditory privacy was a consideration such as in public settings.

\textit{AMIE (Video) was perceived as consistent and thorough:}
Patient actors described AMIE (Video) as providing consistent empathy across encounters and adopting an unhurried conversational approach: \textit{``I did not feel rushed. The AI was thorough and always asked if there was anything else to add''} (P3).
AMIE (Video) was also noted as being \textit{``proactive in wanting to do a physical exam over video''} (P5).
PCPs, however, were preferred for natural conversational cadence and interpersonal connection.

\textit{Perceptual and technical limitations:}
Clinical evaluators documented instances of missed visual cues and uncorrected patient execution errors during guided examination during AMIE (Video) consultations, such as when \textit{``the patient demonstrated the left knee movements and not the right, which was not noted,''}.
Processing silences occasionally caused patient confusion, and evaluators noted specific cases where the agent did not advise on red flag signs.

\subsection{Automated Evaluation Results}
\label{sec:results:automated_evals}

Automated evaluations characterized the contributions of individual agents  to overall system performance and identified remaining capability gaps. 
Full results are provided in \cref{appendix:full_agent_autoeval_ratings,appendix:full_simulated_autoeval_ratings}.

\textit{Specialized agents meaningfully contributed to overall system performance:}
In single-turn evaluations, the full system achieved 59\% compared with 42\% for the Talker-only baseline.
The Perception Agent provided the largest individual contribution, raising overall accuracy to 58\%, with pronounced gains on visual inspection tasks such as \textit{abdominal distension, ascites, and wall contour} (88\% vs 33\%) and \textit{eyelid lesions including chalazia, blepharitis, and ptosis} (91\% vs 56\%).
The Planner Agent complemented these perceptual capabilities through clinical reasoning, improving performance on tasks such as \textit{postural balance loss and gait instability} (54\% vs 32\%) and \textit{joint range of motion and mobility limitations} (65\% vs 50\%).
In multi-turn simulated consultations 
across 20 clinical scenarios from \cite{shah2026towards}, the Planner 
Agent yielded marked improvements in conversational quality and clinical 
thoroughness, with the full system scoring 87\% on overall case-specific rubrics compared with 71\% for the Talker-only ablation.
These quality gains were achieved at substantially lower latency than the architecture described by \citet{shah2026towards}, driven by the transition from sequential to asynchronous agent orchestration, reducing mean turn latency from 21.4s 
[95\% CI: 20.3s, 23.0s] to 2.6s [95\% CI: 2.3s, 2.8s].

\textit{Gaps remained in subtle and high-frequency perceptual competencies:}
Performance in automated evaluations varied considerably across clinical domains. 
Predominantly physical tasks achieved strong results (\textit{Pain behavior}: 88\%, \textit{Dermatologic}: 75\%, \textit{Musculoskeletal}: 70\%), while interactive domains such as \textit{Communication and behavior} (44\%) and \textit{Environment and setup} (46\%) proved more challenging.
Key failure modes involved high-frequency or subtle audio-visual competencies, including \textit{nystagmus} (3\%), \textit{tremors at rest or with action} (24\%), and \textit{affective state, expressiveness, and mood congruence} (29\%).
\section{Related Work}
\label{appendix:related_works}

\subsection{Foundation Models: From Language to Perception}

The development of foundation models for generalist medical artificial intelligence \cite{moor2023foundation} has progressed from static question-answering toward interactive clinical dialogue. Early clinical foundation models demonstrated expert-level medical question-answering on static benchmarks such as MedQA \cite{jin2021disease, Lievin2024-qd, singhal2023large, singhal2025toward, nori2023can, saab2024capabilities}.

However, in medical training, written examinations do not reliably predict clinical competence~\citep{Miller1990-sl, Epstein2002-ps}, as corroborated by the gap between AI performance on static benchmarks and in interactive clinical settings \citep{Schmidgall2024-zf, johri2025evaluation}.
In simulated text-based patient encounters, conversational AI systems have matched or exceeded primary care physicians (PCPs) in diagnostic reasoning, management recommendations, and perceived empathy across multi-turn text interactions  \cite{mcduff2025towards, tu2025towards, lievin2026towards}.
In subspecialist domains, randomized and blinded evaluations showed that AI assistance reduced clinically significant errors among cardiologists in complex cardiovascular care \cite{o2026large} and produced management plans comparable to oncology fellows in breast cancer care \cite{palepu2025exploring}.
Prospective clinical feasibility studies have begun to evaluate conversational agents in real-world ambulatory care, supporting their initial safety and decision-support utility  \cite{brodeur2026prospectiveclinicalfeasibilitystudy, zeltzer2025comparison, korom2025ai}, often under structured multi-agent clinician oversight \cite{mukherjee2024polaris, vedadi2025towards}.

Yet text-based interfaces impose a communicative bottleneck that introduces reporting biases \cite{sackett1979bias} and accessibility barriers \cite{arias2023digital, moore2025patient}, while risking the omission of critical diagnostic cues.
Multimodal clinical models have incorporated static medical imagery and offline data \cite{saab2026advancing, saab2024capabilities, tu2024towards, moor2023med, li2023llava}.
However, active clinical encounters require continuous, real-time audio-visual perception.
While full-duplex speech-to-speech foundation models enable low-latency spoken dialogue \cite{Defossez2024-ie, Xie2024-ui, gdm2026geminiomni}, a disconnect between perception and action persists. Here, current voice systems can reliably identify vocal delivery cues such as distress or sarcasm when queried, but frequently disregard these paralinguistic signals during consequential decision-making  \cite{bartelds2026real}.
Leveraging streaming audio-video infrastructure \cite{GDM_astra}, \citet{shah2026towards} established the feasibility of an audio-visual medical AI across 20 proof-of-concept scenarios demonstrating superiority of a Gemini-based system over a GPT-Realtime baseline, while highlighting important gaps to fill before reaching physician-level performance in this task.
The present work builds on these foundations to provide the first demonstration of expert-level clinical performance in real-time video consultations.

\subsection{Scaling Reasoning with Multi-Agent Architectures}

Expert clinical practice requires balancing rapid conversational rapport with rigorous deliberative reasoning.
In foundation models, scaling test-time compute via chain-of-thought prompting, self-consistency, and search has been shown to substantially improve diagnostic accuracy \cite{wei2022chain, kojima2022large, wang2022self, madaan2023self, singhal2023large, singhal2025toward, snell2024scaling, muennighoff2025s1, gottweis2026accelerating}.
However, while extensive search is well-suited to offline tasks, live clinical consultations impose strict conversational pacing (of only a few seconds of delay between turns), creating a fundamental tension between deep inference-time reasoning and real-time responsiveness.

Multi-agent architectures address this tension by decoupling fast conversational interaction from slower, deliberative reasoning \cite{christakopoulou2024agents, mukherjee2024polaris, lievin2026towards, gottweis2026accelerating}.
Polaris \cite{mukherjee2024polaris} employed specialized clinical agents to monitor conversational safety, while \citet{christakopoulou2024agents} pioneered a dual-agent ``System 1'' (Talker) and ``System 2'' (Reasoner) architecture.
Similar tandem designs have emerged in real-time speech systems to anchor low-latency
spoken dialogue with deliberative knowledge retrieval \cite{kuroki2025kame, su2026multistream}.
Building on the Talker-Reasoner paradigm applied to text-based longitudinal disease management \cite{lievin2026towards}, AMIE (Video) extends this architecture to real-time audio-visual streaming: a 3-agent harness (Talker, Planner, and Perception) that executes deep clinical reasoning and continuous perceptual analysis asynchronously, without interrupting fluid dialogue.

\subsection{Telehealth Examination and Evaluation}

In clinical practice, visual and auditory observations are fundamental aspects of history-taking and physical assessment that cannot be replaced through text \cite{wright2022assessment}.
The growth of telemedicine has established the virtual physical examination as a viable clinical discipline, demonstrating that key maneuvers across musculoskeletal, neurological, dermatological, and respiratory systems can be guided effectively over video \cite{benziger2021telehealth, tong2026translatability, ansary2021virtual, McDonoughUWPhysicalExam, caltrc2021telehealth}.
Moreover, comparative telehealth studies show that video consultations significantly improve diagnostic demonstration and patient satisfaction over audio-only visits \cite{kruis2024patient}.

Objective Structured Clinical Examination (OSCE) \cite{harden1975assessment} remains the gold standard examination format to evaluate clinical competence during medical training, with candidates conducting consultations with professional patient actors, so-called `standardized patients' to complete a series of pre-defined clinical tasks \cite{Gentles2003-ip}.
Medical educators have adapted OSCEs to virtual encounters \cite{sartori2020telehealth}, suggesting rubrics for telehealth communication and diagnostic process.
While prior studies have used OSCE frameworks to evaluate text-based and static multimodal AI \cite{tu2025towards, lievin2026towards, saab2026advancing, vedadi2025towards}, comprehensive assessment benefits from pairing holistic communication scales with granular, case-specific clinical checklists \cite{hodges2003osce}, an approach aligned with contemporary medical dialogue benchmarking \cite{arora2025healthbench, hicks2026healthbench, shah2026towards} and adopted in our video-based OSCE study.
\section{Discussion}
\label{sec:discussion}

Audio-visual capabilities are essential for AI systems to overcome the fundamental diagnostic and communicative bottlenecks of text-only interfaces. Text-based interactions place a heavy burden on patients to translate complex somatic sensations and physical abnormalities into written words, a dependency that often leads to missed cues or ambiguity. In contrast, multimodal audio-video communication unlocks a rich, real-time stream of objective clinical data. It captures nuanced auditory signals, including respiratory effort, vocal
strain, and emotional affect, while enabling direct observation of visual signs such as diaphoresis, scleral icterus, and acute distress. Furthermore, live audio-visual interaction can enable improved clinical rapport and the dynamic guidance of patients through interactive physical examination maneuvers (e.g., abdominal self-palpation, active range of motion, capillary refill) with substantially higher diagnostic fidelity.

The AMIE (Video) system represents a step change from text-bound diagnostic agents to a synchronous, audio-visual interaction. It is the first such system to demonstrate PCP-level performance in a randomized OSCE study setting, as rated by independent clinical evaluators. For many evaluation domains, including case-specific history taking, clinical reasoning, treatment planning, and patient communication, clinical evaluators rated AMIE's performance as significantly better than the baseline PCPs. AMIE (Video) was also preferred for its more proactive approach to physical examination and observation.
Patient actors generally rated AMIE's conversational capabilities favorably compared with those of PCPs, rating AMIE (Video) higher for assessing and explaining their condition.
These higher ratings were achieved within a similar encounter duration (8.9 vs. 9.3 minutes), a notable finding given that longer visit times are strongly linked to increased patient satisfaction and more effective management of chronic and psychosocial conditions \cite{irving2017international, hutton2005benefits}.

These comparative findings are meaningful because the human baseline in this study was designed to closely reflect an actual clinical practice setting. Video-based telemedicine is widely utilized and continues to grow as a standard interaction modality. Prior studies of text-based conversational medical AI \cite{tu2025towards,lievin2026towards,saab2026advancing,vedadi2025towards,brodeur2026prospectiveclinicalfeasibilitystudy} may underestimate physician performance because physicians are more accustomed to telephone or video-based consultation \cite{carrillo2022effectiveness,wharton2019virtual}. This work was designed to be more representative of video-based telemedicine, with PCPs able to see and hear the patient using a familiar video conferencing interface. The PCP baseline in this study therefore serves as a stronger proxy for real-world clinical practice, though key differences remain. Most notably, the study used patient actors instead of actual patients, and physicians were required to keep their cameras off.

Clinical evaluators also rated AMIE (Video) higher than PCPs for its proactivity in guiding physical examinations. In this study, human physicians often leaned on oral history-taking rather than instructing patients to perform active examination maneuvers. This difference may reflect clinicians defaulting to in-person consultation habits, where physical exams are typically performed in the clinic rather than coached remotely. AMIE, by contrast, systematically prompted patients through step-by-step physical examination maneuvers. This behavior highlights a potential opportunity for AI systems, which are less inhibited by traditional time constraints, to elicit more thorough clinical information during virtual care. Indeed, it has been shown that many physical assessment maneuvers can be successfully adapted to synchronous video encounters \cite{benziger2021telehealth, tong2026translatability, ansary2021virtual}. However, further research is needed to understand the clinical necessity and diagnostic yield of such maneuvers relative to the additional time, effort, and cognitive load required from the patient.

The AMIE (Text) arm served as a useful ablation to characterize the impact of audio-visual interaction. Patient actors significantly preferred the audio-visual communication for its ease of use, communicative effectiveness, and ability to make the user of the system feel understood. Clinical evaluators rated AMIE (Video) as superior to AMIE (Text) in several domains key to the telehealth setting, such as fostering a relationship, audio-visual perception, and guided examination, as well as overall across all five case-specific clinical rubrics in aggregate.
However, for rubrics related to clinical reasoning such as diagnostic and management appropriateness, AMIE (Video) was comparable to AMIE (Text). 
In part, this may reflect a trend towards approaching saturation on these simulated OSCE scenarios, where text-based clinical reasoning performance is already high. This finding also aligns with clinical literature suggesting that the medical history alone establishes the correct diagnosis in the vast majority of outpatient encounters, while physical examination alters the primary diagnosis in only a small fraction of cases \cite{hampton1975relative, peterson1992contributions}. Similar results have been observed in other recent telehealth studies, showing comparable clinical effectiveness between telephone and video consultations \cite{byambasuren2023comparison}, though other studies have identified reductions in medication errors and improved diagnostic accuracy \cite{rush2018videoconference} with the addition of video. Because a missed visual cue can often be compensated by verbal questioning, the distinct advantages of audio-visual reasoning may only emerge in specific situations, such as when a patient’s spoken statements conflict with visual evidence (e.g., verbally naming one medication while holding up a different pill bottle). These simulated scenarios were designed to reflect a breadth of telehealth settings including opportunities for physical observation and examination. However, they were not designed to render purely text-based consultations impossible, which helps to contextualize this finding. 

The granular, case-specific rubrics (\cref{sec:case_specific_rubrics}) were designed to more precisely detect nuanced variations in clinical performance and communication that the broader general rubrics may overlook.
As clinical agents improve and metrics like diagnostic accuracy become more saturated, these targeted scoring criteria, which can be tailored to the precise challenges of each scenario, may help uncover more subtle differences in clinical competency.
This approach mirrors recent shifts in medical LLM evaluation, such as HealthBench \cite{arora2025healthbench,hicks2026healthbench}, which evaluates open-ended responses to clinical dialogues against physician-authored conversation-specific rubrics. 
In OSCE examinations used in medical education, station-specific checklists are often utilized \cite{harden1975assessment}; combining general clinical rubrics with case-specific criteria ensures a balanced and standardized assessment of the overall conversational appropriateness and clinical reasoning \cite{hodges2003osce}. 

The real-time delivery of high-quality dialogue from AMIE was enabled by its decoupled, asynchronous multi-agent architecture. This design addresses a fundamental tension in real-time medical dialogue: the need for near-instantaneous conversational turn-taking (low latency) versus the computational expense of thorough diagnostic reasoning and continuous visual processing.
Automated evaluation ablations confirm that this three-agent harness substantially outperforms the single-agent baseline and prior harnesses from \citet{shah2026towards}, while operating at a natural conversational latency (around 3 seconds per turn).
These ablations also show how individual components of AMIE (Video) meaningfully contribute to overall system performance.

A rigorous offline evaluation framework is essential for systematically benchmarking agent performance and tracking progress throughout development. Our evaluation framework comprises two components: the clinical audio-visual taxonomy (\cref{appendix:av_cues}) and a multi-tiered benchmark suite. Grounded in telehealth feasibility, the taxonomy defines a principled spectrum of audio-visual signs that can be elicited and interpreted in virtual care. While exhaustively covering every clinical action and condition in the taxonomy was beyond the scope of this work, our approach establishes an extensible blueprint for evaluating multimodal clinical AI. Specifically, we pair complementary single-turn benchmarks, which isolate and evaluate discrete instances of clinical perception and diagnostic reasoning, with multi-turn simulated consultations that assess holistic, conversation-level competencies. Together, these tiers provide a computationally and logistically scalable proxy for the evaluation of real-time clinical audio-video consultations. Although the present multi-turn audio simulation communicates visual cues descriptively, future evaluation environments can directly incorporate dynamically generated synthetic patient videos \cite{nasimzada2024towards}.

While the agents in AMIE (Video) are orchestrated through prompting, multi-turn reinforcement learning with simulated patients has been shown to yield robust improvements in diagnostic accuracy and clinical safety \cite{lievin2026residencyrl}. The simulated encounters and structured evaluation criteria developed in this work could serve as a training environment for such end-to-end optimization.

\subsection{Limitations}
The system described in this work is not ready for real-world clinical deployment and several core limitations must be acknowledged.
An important architectural constraint of AMIE (Video) is its reliance on discrete turn-taking rather than continuous, multi-stream interactivity. Although the decoupled multi-agent design balances conversational latency and reasoning by running background agents asynchronously, the ultimate interaction loop still hinges on more rigid turn boundaries. Consequently, the patient and agent cannot speak simultaneously or engage in rapid, overlapping back-and-forth dialogue as human clinicians and patients naturally do during consultations. Furthermore, this audio-based separation also prevents visually triggered interjections, such as speaking up to guide a patient actor who is improperly executing a physical examination maneuver.
Recent paradigms, such as interaction models \cite{thinkingmachines2026interaction}, multi-stream LLMs \cite{su2026multistream}, and tandem speech architectures like KAME \cite{kuroki2025kame}, address these sequential bottlenecks by enabling full-duplex communication (``speaking while 
listening'').

Despite the strong end-to-end performance AMIE (Video) demonstrated in our OSCE evaluation study, automated evaluations uncover important capability gaps that warrant careful examination. Subtle affective nuances and high-frequency movements like tremor remain difficult for the agent to reliably perceive.
These granular perceptual capabilities are crucial to clinical reasoning for some medical presentations and therefore should be further developed in future work to increase robustness and coverage of audio-visual perceptive capabilities.  
Additionally, even when an agent successfully perceives subtle non-verbal cues, it may still fail to translate this perception into appropriate clinical action. During automated evaluations, we observed instances where the agent's perception was accurate, yet did not trigger the appropriate proactive response (such as noticing poor audio quality but failing to request a correction). This mirrors other recent findings which demonstrate that voice agents can recognize properties like distress or sarcasm but fall short of fully incorporating them in their decision-making \cite{bartelds2026real}.

Constructing and executing automated evaluations for complex interactive domains like audio-visual clinical consultations is a process with inherent challenges. Patient-acted videos exhibit varying degrees of clinical realism, while internet-sourced videos can lack the focused and natural presentation typical of live telehealth encounters. Furthermore, evaluating perception, reasoning, and interactivity in a targeted single-turn manner involved designing occasionally contrived examples (e.g., a patient directly asking, “It hurts right here... what part is this?”) to strictly isolate the capability being tested. Additionally, while we included a limited number of `adversarial' examples, future iterations of the framework may enrich this adversarial subset with additional scenarios, such as patient actors performing examination maneuvers incorrectly or contradicting physical observations. During automated evaluation, the patient simulator's behavior can vary slightly across simulation runs, and auto-graders exhibit a certain degree of intra-rater variability. Therefore, performance comparisons stemming from this automated process should be interpreted with appropriate caution. While automated evaluation results are valuable for relative comparisons of multiple candidate systems, their calibration with clinical evaluator ratings may not be precise. Consequently, these automated evaluations were primarily used to assess directional improvements of the system as sub-components were added or improved. However, they are not a suitable replacement for rigorous randomized studies involving human physician comparisons, professional patient actors and evaluation ratings from independent clinical evaluators.

The clinical scope of the scenarios selected for the OSCE study were limited. Because evaluations relied entirely on live enactments by professional patient actors, the 100 study scenarios were restricted to ``actable'' clinical presentations. Consequently, important clinical scenarios featuring core audio-visual diagnostic components such as dermatological conditions that cannot be authentically simulated by actors were omitted from the study. Furthermore, while patient actors provide a useful standardized baseline for comparative evaluation, real-world clinical encounters introduce unscripted behavioral, environmental, and emotional complexities.

A related constraint involved demographic alignment between patient actors and the simulated clinical conditions. While experimental protocols successfully matched actor biological sex to the clinical case descriptions and maintained a broad distribution across racial backgrounds and age groups, this challenge was particularly pronounced in simulating complex chronic and acute illnesses that disproportionately affect geriatric populations (e.g., severe cardiovascular disease, chronic obstructive pulmonary disease). This skew reflects logistical constraints in recruiting older standardized patients as well as the digital literacy required for synchronous teleconferencing platforms which remains a documented barrier to telemedicine access among older populations \cite{lam2020assessing, o2022overcoming, mao2022barriers}. Evaluating conversational AI with older patient cohorts who present with multi-morbidity and sensory or cognitive impairments is an essential next step for clinical generalizability.

While our experimental protocol implemented several measures to standardize conditions between study arms, complete blinding remained challenging.
Standardization measures included utilizing an identical teleconferencing interface (capable of video interaction and text chat modalities) across all arms, asking PCPs to turn off their own camera during consultations to match the avatar-less AI system, and randomizing scenario presentation order and evaluation sequence across arms. 
Limitations to blinding include the synthetic vocal style of the Talker agent which remains audibly distinct from human physician speech, and the fact that the text-only arm represented an inherently different interaction format.
Such unblinding is a critical consideration, as awareness of an agent’s AI identity can directly affect user behavior; it was observed in~\citet{reis2026reduced} that patients who believe they are interacting with an AI chatbot rather than a physician demonstrate lower-quality symptom reports.
Furthermore, asking PCPs to disable their cameras (a decision made to eliminate visual appearance confounders) reduced the ecological validity of the human baseline. Seeing the physician's facial expressions, eye contact, and non-verbal presence contributes meaningfully to patient trust, empathy, and interpersonal rapport \cite{zulman2020practices, hall1995nonverbal, mast2007importance}.

Finally, our evaluation study tests AMIE in the setting of video-based telehealth.
While this is a common modality routinely used for remote clinical care, there are inherent clinical limits to this modality.
Virtual clinical agents and telehealth practitioners alike are unable to perform certain physical examinations over a telehealth interface, such as palpating a spleen, assessing deep abdominal rigidity, or performing fundoscopy. 
Furthermore, discerning subtle audio-visual signs may be challenging in a telehealth setting, especially when audio or video quality is poor.

\section{Conclusion}
\label{sec:conclusion}

This research advances AMIE towards real-time clinical video consultations, providing the first demonstration of expert-level AI performance in this modality through a randomized OSCE study with board-certified PCPs and professional patient actors across 100 clinical scenarios. Unlike prior text-chat evaluations, this comparison reflects the modality that physicians and patients routinely use for telehealth, offering a more ecologically valid assessment of clinical performance. These results were enabled through a novel asynchronous multi-agent architecture balancing low-latency dialogue with thorough clinical reasoning and real-time audio-visual perception. While AMIE matched or exceeded PCPs across core clinical competencies in this video configuration, PCPs were preferred for rapport and partnership building, a reminder that clinical interactions encompass dimensions beyond diagnostic and management accuracy. Limitations remain in fine anatomical precision, affective nuance, and high-frequency movement perception, and the study relied on patient actors rather than real patients. Substantial validation in real clinical workflows is essential before responsible translation. By engaging with the perceptual richness of clinical practice, this work represents a step toward AI systems that can more equitably expand access to high-quality care.

\subsubsection*{Acknowledgments}
This project was an extensive collaboration between many teams at Google Research and Google DeepMind. 
We thank Rory Sayres and Yishay Mansour for their comprehensive review and detailed feedback on the manuscript. 
We also thank Jacinda Mein, Jen Haslip, Amira Fouad, Evie Gray, Danielle Breen, Amanda Carl-Pratt, Isha Mishra, Edna Agyemang, Aziza Henry, Kate Mischaikow, Louis Jardin, Kevin Soares, Esmeralda Cardenas, and Tim Herrmann for contributions to the narratives and visuals. 
We are grateful to Rachelle Sico, Cat Kozlowski, Jessica M. Williams, Celeste Grade, Oba Adewunmi, and Michael Cael Davis for their support during our research.
We are grateful to Alan Karthikesalingam, Jim Taylor, John Hernandez, Naama Hammel, Susan Thomas, Raia Hadsell, Zoubin Ghahramani, Greg Corrado, Pushmeet Kohli, Michael Howell, and Demis Hassabis for their valuable support, insights and feedback during our research.
We also thank Medical Science Solutions Inc. and Christopher Smith for their partnership in conducting the OSCE study, as well as SiWai Man, Brett Hatfield, Gordon Turner and Priyana Acharya.
Finally, we are grateful for the support from Adam Rodman, Jason Gusdorf, Daniel Cook, Priya Gupta, Jeremy Lai, Jon Krause, Jonas Kemp, Kejia Chen, Alexey Kolesnikov, Pi-Chuan Chang, Qinghan Xue, Kishwar Shafin, Theo Guidroz, Fan Zhang, Diego Ardila, Lucas Brambrink, Quang Duong, Yuexing Hao, Bijal Rajput, Stefan Moser, Niko Schwarz, Alex Bijamov, Ayush Jain, Bhavna Daryani, Cara Tan, Daniel McDuff, Gal Elidan, Hieu Hoang, Ines Mezerreg, Jackie Barr, Jacqueline Shreibati, Joe Breda, Laura Vardoulakis, Lin Yang, Mike Sanchez, Preeti Singh, Tiya Tiyasirichokchai, Will Vaughan, Xiang Ji, Sara Mahdavi, David Barrett, David Stutz and Aleksandra Faust.

\subsubsection*{Data Availability}
Example evaluation scenarios and excerpts of patient actor scenario enactments are provided in the manuscript and appendix.
For privacy reasons, we do not release recordings of patient actors at this time.

\subsubsection*{Code Availability}
AMIE is an LLM based research AI system. We are not open-sourcing model code and weights due to the safety implications of unmonitored use of such a system in medical settings. In the interest of responsible innovation, we will be working with research partners, regulators, and providers to validate and explore safe onward uses of AMIE. For reproducibility, we have documented technical methods while keeping the paper accessible to a clinical and general scientific audience. Our work builds upon Gemini 3.0 Flash and Gemini 3.1 Pro, with details available in the corresponding model cards \cite{gemini3flash,gemini3p1pro}.

\subsubsection*{Competing Interests}
This study was funded by Alphabet Inc and/or a subsidiary thereof (`Alphabet'). Authors who are employees of Alphabet may own stock as part of the standard compensation package.

\newpage
\setlength\bibitemsep{3pt}
\printbibliography
\clearpage
\end{refsection}

\begin{refsection}
\clearpage

\renewcommand{\thesection}{A.\arabic{section}}
\renewcommand{\thefigure}{A.\arabic{figure}}
\renewcommand{\thetable}{A.\arabic{table}} 
\renewcommand{\theequation}{A.\arabic{equation}} 
\renewcommand{\theHsection}{A\arabic{section}}

\setcounter{section}{0}
\setcounter{figure}{0}
\setcounter{table}{0}
\setcounter{equation}{0}

\noindent \textbf{\LARGE{Appendix}}\\
\normalfont

\begin{enumerate}
    \item \cref{appendix:av_cues} describes the taxonomy of audio-visual cues and examination maneuvers that was used to scope auto-evaluation.
    \item \cref{appendix:agent_eval_design} provides details on the development and mechanics of single-turn auto-evaluation examples, which were used to develop the system and characterize its clinical audio-visual capabilities.
    \item \cref{appendix:full_agent_autoeval_ratings} provides detailed results on the single-turn auto-evaluations performed on different versions of the AMIE system.
    \item \cref{appendix:full_simulated_autoeval_ratings} provides detailed results on the simulated auto-evaluations performed on different versions of the AMIE system.
    \item \cref{appendix:pilot_run} describes the 20-scenario pilot study conducted across 6 experimental arms to validate the system architecture and familiarize participants and includes automated evaluation results.
    \item \cref{appendix:osce_study_details} describes further details on the 100-scenario OSCE study with 3 study arms and subsequent clinical evaluation.
    \item \cref{appendix:case_specific_rubric_example} includes a full example of a scenario, ground truth, and corresponding case-specific rubric.
    \item \cref{appendix:av_telehealth_rubric} lists the supplementary evaluation rubric adapted from \citet{sartori2020telehealth} to assess telehealth audio-visual symptom observation and guided physical examinations.
    \item \cref{appendix:post_questionnaire} details the structured post-encounter clinical questionnaire completed by PCPs and AV-AMIE following each consultation.
    \item \cref{appendix_section:patient_actor_ui} shows the UIs used by AMIE and PCPs across different study arms, as well as the interface used by expert raters.
    \item \cref{appendix:tabular_results} contains detailed results for the OSCE study in a tabular format.
    \item \cref{appendix:full_patient_actor_ratings} presents detailed ratings from patient actors.
    \item \cref{appendix:full_expert_eval_ratings} presents detailed ratings from the expert evaluators.
    \item \cref{appendix:patient_actor_interviews} presents details on structured interviews with patient actors.
\end{enumerate}

\clearpage
\section{Taxonomy of audio-visual cues and physical exam maneuvers}
\label{appendix:av_cues}

To establish an audio-visual telehealth evaluation framework, clinically relevant observations and patient-assisted examination maneuvers were organized into the taxonomy shown in \cref{table:av_cues_taxonomy}. The taxonomy separates four categories of information: (1) passive audio-visual observations, (2) patient-assisted examination maneuvers, (3) interaction behaviors that improve data quality, and (4) explicit boundaries of telehealth assessment. This distinction is important because certain clinically meaningful cues may be challenging to elicit in a virtual setting. Accordingly, each capability was annotated by telehealth feasibility (\cref{table:av_cues_feasibility_legend}).

A key feature of this taxonomy is distinguishing findings that can be accurately simulated by standardized patients from those that require the presence of a clinical condition or abnormality or can only be elicited by in-person examination. Standardized patients are well suited for some functional, behavioral, and movement-based cues. In contrast, physical signs such as skin lesions, examination of orifices, or those requiring medical equipment (e.g. auscultation) or close up images generally require clinician-reviewed images or videos, high-quality close-up visualization, peripheral devices, or in-person assessment. For this reason, the OSCE study emphasized actable scenarios (\cref{sec:osce_study}), whereas the automated evaluations incorporated both actable cases and clinician-reviewed curated media (\cref{sec:methods:agent_evaluation,sec:methods:multiturn_evaluation}) incorporating physiologic abnormalities.

The taxonomy was developed by board-certified family medicine and internal medicine clinicians, who synthesized physical-examination benchmarks from the University of Washington School of Medicine \cite{McDonoughUWPhysicalExam}, the Caravan Health telehealth physical-examination reference guide \cite{caltrc2021telehealth}, the virtual physical examination framework described by Ansary et~al. \cite{ansary2021virtual}, and textbook sources \cite{dover2023macleod}. These sources were consolidated into a multi-system inventory of cue families, feasibility levels, and benchmark references.

\begin{table}[h!]
\centering
\footnotesize
\renewcommand{\arraystretch}{1.15}
\caption{\textbf{Telehealth-feasibility categories used in the audio-visual cue taxonomy.}}
\label{table:av_cues_feasibility_legend}
\begin{tabularx}{\textwidth}{@{}>{\RaggedRight\arraybackslash}p{1.4cm}>{\RaggedRight\arraybackslash}p{3.2cm}>{\RaggedRight\arraybackslash}X@{}}
\toprule
\textbf{Code} & \textbf{Interpretation} & \textbf{Evaluation implication} \\
\midrule
\textbf{H} & High feasibility & The finding or maneuver can usually be observed, elicited, or derived from history taking through standard audio-video with minimal setup. \\
\addlinespace[2pt]
\textbf{M} & Moderate feasibility & The finding may be assessable only with favorable lighting, camera position, audio quality, patient cooperation, or patient-assisted examination. Responses should explicitly communicate uncertainty when the signal is limited. \\
\addlinespace[2pt]
\textbf{L} & Low feasibility & The finding is unreliable through standard audio-video alone and generally requires high-quality close-up imaging, a trained helper, specialized peripheral equipment, or in-person examination. These cases primarily evaluate caution and boundary recognition. \\
\addlinespace[2pt]
\textbf{N} & Not feasible & The finding cannot be adequately assessed through standard audio-video alone. The expected behavior is to avoid claiming that the finding was assessed and to recommend in-person or device-assisted evaluation when clinically relevant. \\
\bottomrule
\end{tabularx}
\end{table}

\begin{table*}[!t]
\centering
\caption{\textbf{Taxonomy of audio-visual clinical cues and patient-assisted examination maneuvers for telehealth evaluation.}
Telehealth-feasibility codes are defined in \cref{table:av_cues_feasibility_legend}.
A/V = audio-visual; obs. = observation; pt. = patient; sim. = simulated; JVD = jugular venous distention.}
\label{table:av_cues_taxonomy}
\vspace{0.3em}

\begingroup
\tiny
\renewcommand{\arraystretch}{1.4}
\setlength{\tabcolsep}{3.5pt}
\hyphenpenalty=10000
\exhyphenpenalty=10000
\emergencystretch=1em

\begin{tabularx}{\textwidth}{@{}
>{\raggedright\arraybackslash}p{2.00cm}
>{\raggedright\arraybackslash}p{4.50cm}
>{\raggedright\arraybackslash}p{1.50cm}
>{\centering\arraybackslash}p{0.70cm}
>{\raggedright\arraybackslash}p{2.00cm}
>{\raggedright\arraybackslash}X
@{}}
\toprule
\textbf{Area} &
\textbf{Cues or maneuvers} &
\textbf{Acquisition} &
\mbox{\textbf{TH-F}} &
\textbf{Source} &
\textbf{Evaluation focus and feasibility caveats} \\
\midrule

\textbf{Abdominal}
& Visible discomfort, guarding, distension, pain localization, self-palpation response, movement-associated pain
& Guided self-exam
& M
& Actor; sim.; OSCE
& Coaches safe self-palpation, localizes tenderness, and escalates severe or focal pain. Deep palpation, rebound tenderness, rigidity, organomegaly, fluid wave, and shifting dullness are low-feasibility. \\

\textbf{Cardiovascular}
& Home pulse/BP, capillary refill, visible edema, cyanosis, pallor
& Pt. measurement; visual obs.; guided maneuver
& \mbox{M--L}
& Actor; sim.; media; OSCE
& Uses patient-reported and visible perfusion findings cautiously. Cyanosis, JVD, pitting edema, and perfusion signs are lower-feasability on standard video. \\

\textbf{Communication and behavior}
& Affect, anxiety, agitation, psychomotor slowing, eye contact, engagement, speech cadence
& Passive A/V obs.
& H
& Actor; OSCE
& Integrates behavioral and paraverbal cues into rapport, mental status, and severity assessment without over-interpreting normal, cultural, or neurodiverse variation. \\

\textbf{Constitutional}
& General appearance, distress, fatigue, diaphoresis, pallor, apparent discomfort
& Passive visual obs.
& \mbox{H--M}
& Actor; media; OSCE
& Recognizes acuity and distress. Pallor, diaphoresis, and color change are lighting- and skin-tone-dependent and require caution. \\

\textbf{Dermatologic}
& Rash/lesion pattern, distribution, swelling, bruising, wounds, ulcers, mottling
& Visual obs.; guided close-up view
& \mbox{M--L}
& Images/clips; OSCE
& Describes visible morphology and distribution. Color, texture, depth, warmth, tenderness, and subtle lesions depend on image quality; avoid definitive diagnosis with poor quality. \\

\textbf{Environment and setup}
& Lighting, camera angle, audio quality, framing of relevant body region
& Data-quality interaction
& H
& Actor; OSCE
& Requests improved lighting, closer camera positioning, clearer audio, or safer framing before interpreting visual findings. \\

\textbf{HEENT}
& Conjunctival injection, facial swelling, nasal congestion, voice change, oral/throat view, extraocular movements, facial symmetry
& Observation; guided camera; guided maneuver
& M
& Actor; media; OSCE
& Guides camera use safely and recognizes gross abnormalities. Pupil reactivity, red reflex, views of the oropharynx, mucosa, require close-up imaging or devices. \\

\textbf{Musculoskeletal}
& Active range of motion, restrictions or discomfort with movement, posture and gait abnormalities, joint or limb swelling or redness or asymmetry, self-palpation tenderness
& Guided self-exam
& \mbox{H--M}
& Actor; OSCE
& Coaches movement safely and interprets functional limitation. Strength testing, joint stability from maneuvers, warmth, effusion, and fine deformity are lower-feasibility. \\

\textbf{Neurologic}
& Slurred speech, confusion, altered attention, facial asymmetry, pronator drift, tremor, gait abnormality, coordination, gross strength asymmetry
& Passive A/V obs.; guided maneuver; questions
& \mbox{H--M}
& Actor; media; OSCE
& Recognizes acute neurologic red flags, instruction clarity, laterality, and asymmetry. Reflexes, tone, subtle weakness, sensory testing, and detailed cognitive testing are lower-feasibility. \\

\textbf{Pain behavior}
& Wincing, splinting,  reluctance to move, abnormal posture of body parts, guarding, visible discomfort during maneuvers or gait
& Visual obs.; maneuver response
& \mbox{H--M}
& Actor; OSCE
& Uses non-verbal pain behavior to guide follow-up and management; it should complement, not replace, patient-reported symptoms. \\

\textbf{Respiratory}
& Cough, hoarseness, inspiratory (stridor) or expiratory (wheeze) sounds, dyspnea while speaking, increased work of breathing, accessory muscle use, pursed lip breathing, cyanosis.  
& Passive A/V obs.
& \mbox{H--M}
& Actor; media; OSCE
& Detects respiratory difficulty and escalates appropriately. Wheeze, stridor, significant rapid breathing, pursed lip breathing, cyanosis all lower feasibility. Capturing subtle work of breathing depends on audio quality, framing, and severity. \\

\textbf{Safety and boundaries}
& Severe dyspnea, chest pain with distress, acute neurologic deficit, fall risk, unsafe self-exam conditions, privacy-sensitive exposure, self harm
& Observation; interaction control
& H
& Actor; sim.; OSCE
& Stops unsafe maneuvers, avoids privacy-invasive requests, states telehealth limitations, and recommends urgent or in-person care when indicated. \\

\textbf{Spatial orientation and localization}
& Pointing gestures, anatomical landmark identification, body part recognition, laterality
& Passive visual obs.
& H
& Actor; media; OSCE
& Maps pointing gestures to the correct anatomical landmarks. Correctly distinguishes the patient's left/right from the viewer's perspective, regardless of patient position.\\

\textbf{Inaccessible by A/V}
& Auscultation,  percussion, deep abdominal palpation and other body parts (e.g. breast, genitals, back, etc), detailed ear or eye examination (fundoscopy, otoscopy) w/o device, pulse oxygenation w/o device, genital/rectal exam, other intimate exam
& Boundary condition
& N
& Boundary tests
& Avoids fabricating findings and states that assessment requires in-person examination, a trained examiner, or appropriate peripheral devices. \\

\bottomrule
\end{tabularx}
\endgroup
\end{table*}

\clearpage
\section{Single-Turn Agent Auto-evaluation Design}
\label{appendix:agent_eval_design}

We used a multi-stage curation pipeline to  construct evaluation scenarios (\cref{figure:agent_eval_rubric}) from the clinical taxonomy in \cref{appendix:av_cues}:
\begin{itemize}
    \item \textbf{Enumeration of audio-visual cues:} We started by enumerating the list of audio-visual cues described in \cref{appendix:av_cues} and separating them based on whether they were ``actable'' or not.
    \item \textbf{Actable Scenario Generation:} For ``actable'' clinical cues, an LLM (Gemini 3 Flash) was used to generate multiple draft scenarios. Each scenario establishes a short dialogue scene culminating in a specific ``target turn.'' (for example, a patient abruptly coughing, with the agent evaluated on its capacity to recognize this in either its verbal response or its internal reasoning traces). An LLM classifier along with human clinicians further filtered to the scenes that were actable. One of 24 team members enacted the remaining scenes.
    \item \textbf{Real-World Video Sourcing:} Non-actable audio-visual cues from \cref{appendix:av_cues} were sourced from videos on YouTube and Reddit where possible. From each source artifact, a time window isolating the relevant sign was extracted. To reliably elicit an agent response, synthetic conversation history was integrated where necessary without being suggestive of the particular audio-visual sign being tested. For example, a video depicting a user with conjunctival erythema was augmented with a synthetic audio prompt of "Doctor, what do you notice?" 
\end{itemize}

Grading instructions for each scenario were generated based on the scenario details and transcript. These scenario-specific grading instructions underwent human clinician refinement for a random subset of 200 cases, with the rest going through automated review. In total, the finalized evaluation suite comprises 217 (43.4\%) patient-acted cases and 283 (56.6\%) examples curated from real-world patient videos, yielding a benchmark of 500 single-turn evaluation scenarios. The final evaluation set covers many, but not all of the capabilities described in the taxonomy.

\begin{figure}[h!]
    \centering
    \includegraphics[width=\textwidth]{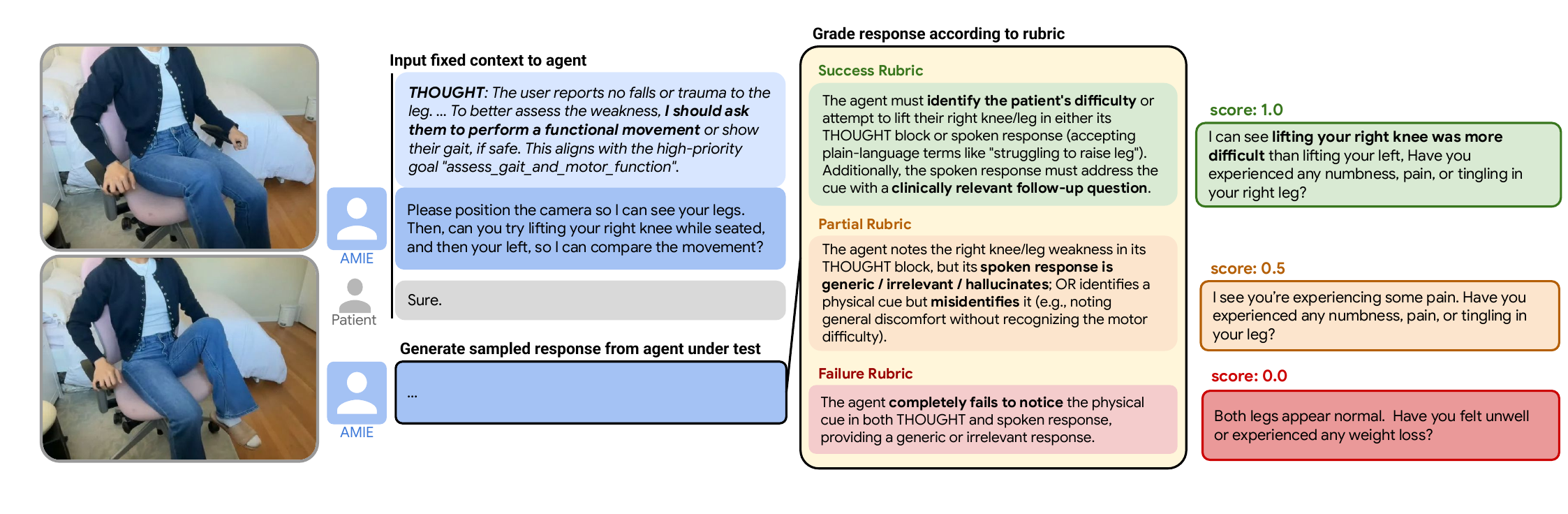}
    \vspace{0.1cm}
    \caption{\textbf{Single-turn Agent Evaluation Example.} Each single-turn agent evaluation serves as a unit test for telehealth capabilities. In this illustrative example, the agent is expected to both perceive a clinical cue, i.e. the user's motor difficulty, and reason about it to propose a clinically relevant follow-up question, e.g. further assessing the user's gait and motor function or their experience of the pain. In evaluation, the previous conversation history is provided as fixed context, then the agent response is sampled at test. Next, a Gemini 3 Flash-based auto-rater scores the response based on a rubric, which has been auto-generated with clinical guidance on the ground truth.}
    \label{figure:agent_eval_rubric}
\end{figure}

\clearpage
\section{Single-Turn Agent Auto-evaluation Ratings}
\label{appendix:full_agent_autoeval_ratings}

The 500 single-turn agent evaluation examples, described in \cref{sec:methods:agent_evaluation,appendix:agent_eval_design}, were used to compare different configurations of the AMIE (Video) system, including the full system, a version without the Perception Agent, a version without the Planner Agent, and a version without both of these background agents. Furthermore, the agent harness presented in \citet{shah2026towards} was tested (updating the model endpoints from Gemini 2.5 Flash to Gemini 3.0 Flash for better comparability), as well as a version of that system without their sequential clinical Planner agent. Gemini 3.0 Flash served as the LLM-based autorater for these evaluations, and each example was tested four times per configuration. To aid in visualization and interpretation, we used Gemini 3.0 Flash to map each evaluation example to group related examples into discrete audio-visual competency descriptions, based on their transcript and ground truth.

\begin{figure}[h!]
    \centering
    \includegraphics[width=\textwidth]{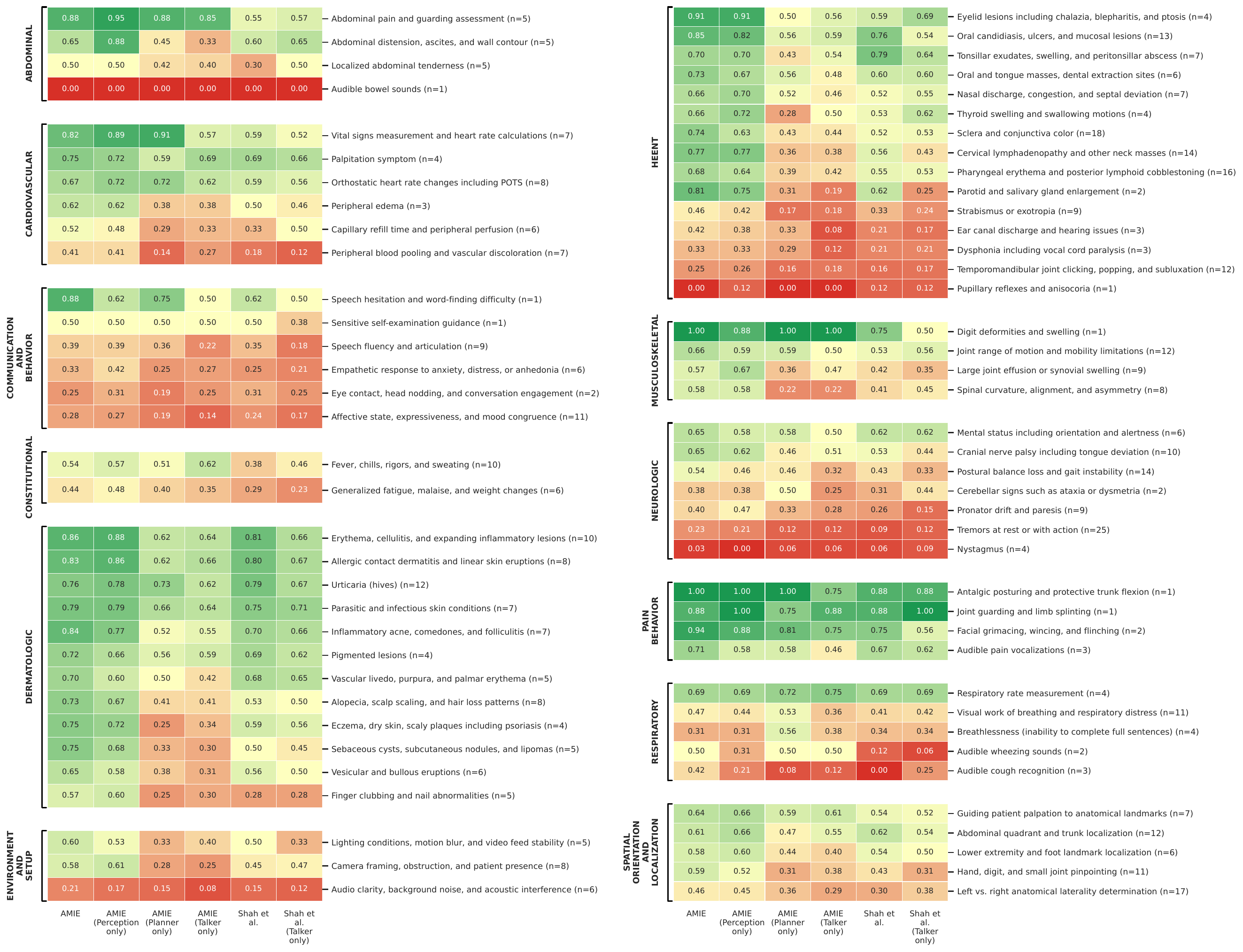}
    \vspace{0.1cm}
    \caption{\textbf{Agent Evaluation Ratings.} Average auto-rated score across 4 replications per agent configuration on audio-visual evaluation examples. For visualization purposes, examples are grouped into low-level descriptions as well as high-level taxonomy categories from \cref{appendix:av_cues}.}
    \label{figure:full_agent_eval}
\end{figure}
\clearpage
\section{Simulated Auto-evaluation Ratings}
\label{appendix:full_simulated_autoeval_ratings}

The 20 scenarios from \citet{shah2026towards} were simulated with an LLM-based patient simulator and LLM-based autorater both using Gemini 3 Flash. Notably, these scenarios were also designed with case-specific rubrics, although the criteria had been organized into 7 domains rather than the 5 domains used in the primary OSCE study (\cref{appendix:osce_study_details}). These simulated evaluations were used while developing the eventual AMIE (Video) system to guide prompting and system design; improving latency, proactivity of physical observation and examination, and completeness of history taking were major areas of focus. 

Here, we report the performance of various versions of the AMIE (Video) system, including the full system, the system without the Perception Agent, without the Planner Agent, and without both of these background agents. Additionally, the agent harness presented in \citet{shah2026towards} was tested (updating the model endpoints from Gemini 2.5 Flash to Gemini 3.0 Flash for better comparability) with and without their sequential clinical Planner agent. We present auto-evaluated scores on the case-specific rubrics from these scenarios, as well as the mean response latency for each agent during the simulation.

\begin{figure}[h!]
    \centering
    \includegraphics[width=\textwidth]{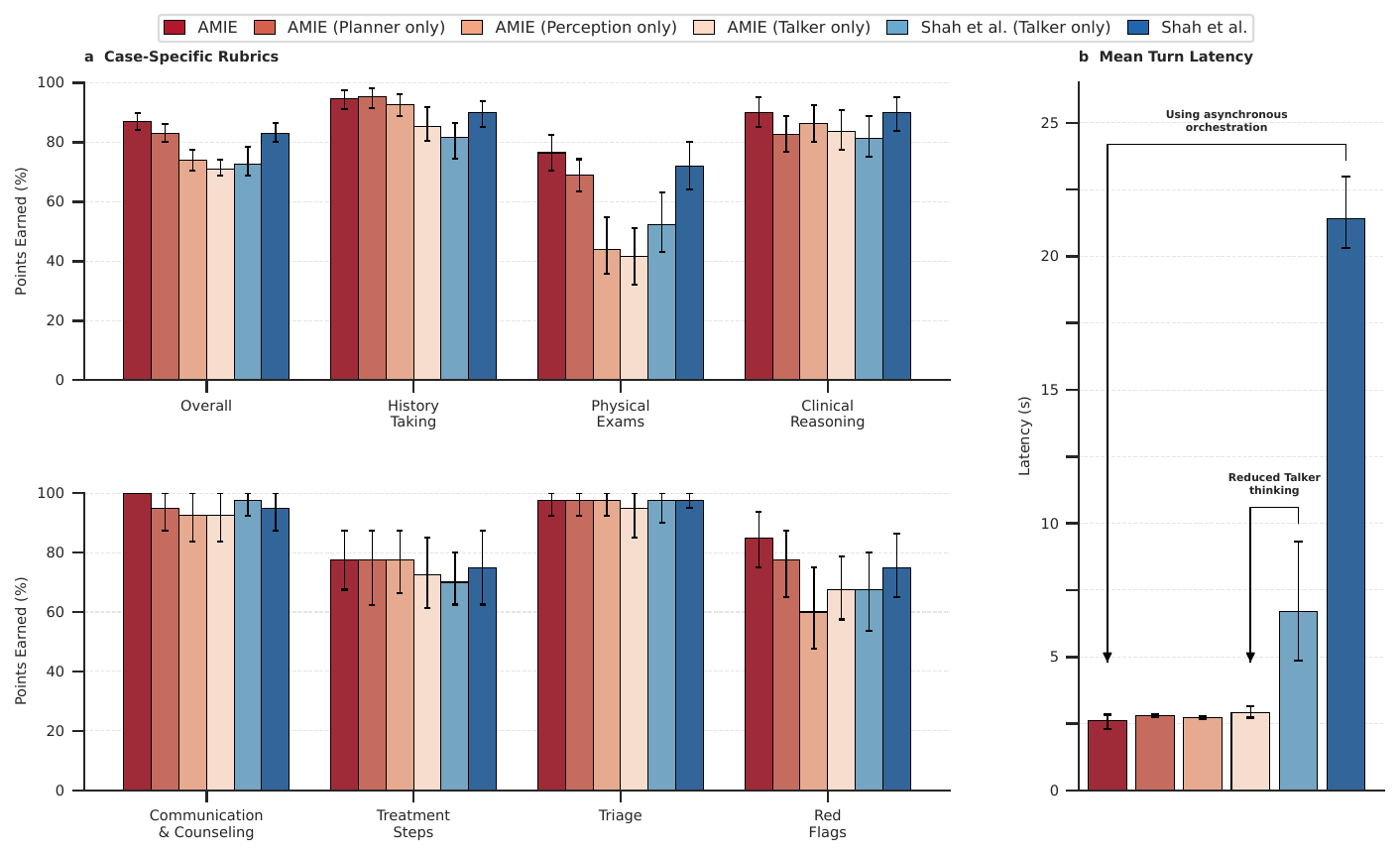}
    \vspace{0.1cm}
    \caption{\textbf{Auto-evaluation Results from Simulated Multi-turn Conversations.} Conversations were simulated between an audio-based patient grounded on one of the twenty scenarios from \citet{shah2026towards}. a) Auto-evaluated ratings on the case-specific rubrics from these scenarios, spanning 7 clinical domains as well as overall. b) Mean latency of the different agents. Error bars correspond to 95\% bootstrap confidence intervals (n=1000).}
    \label{figure:full_simulated_autoeval_ratings}
\end{figure}
\clearpage
\section{Pilot Study: System Validation and Modality Ablation}
\label{appendix:pilot_run}

Prior to executing the main 100-scenario clinical evaluation study (\cref{sec:osce_study}), we conducted a dedicated pilot study consisting of 20 clinical scenario packs. This preparatory phase served three fundamental operational and scientific objectives: (1) verifying the end-to-end reliability of the real-time audio-visual interface, (2) familiarizing professional patient actors and primary care physicians (PCPs) with the study interfaces and interaction dynamics, and (3) performing a small-scale controlled, multi-modal ablation for both AMIE and PCPs to understand the contributions of audio alone versus full synchronous audio-video interaction.

\subsection{Study Design and Automated Evaluation Methodology}

The pilot study evaluated 20 scenario packs, which had the same form as those used in the main study (see \cref{appendix:case_specific_rubric_example}). The study design was largely similar to the main study, except each simulated patient interacted with six arms instead of three. The six arms in this pilot run comprised of both AMIE and baseline PCPs across three communication modalities: text-only, audio-only, and synchronous video+audio. The same patient actor interacted with each arm. A different PCP was used for the three PCP sessions for a given scenario to eliminate the potential of the PCPs using information from prior interactions. During this pilot stage, human expert evaluators were not utilized; instead, consultations were evaluated with an auto-rater (LLM-as-a-judge). Given the consultation transcript and post-questionnaire responses, Gemini 3.0 Flash generated scores for each criterion in the case-specific rubric (20 points total).

\subsection{Pilot Study Results}
\label{appendix:pilot_results}

The automated evaluation results suggest that AMIE is able to have clinically meaningful conversations (rated at or above those of PCPs) with the patient actors in all modality settings.  AMIE and PCPs benefit from access to both audio inputs and video inputs. The automated evaluation metrics across the six arms are summarized in \cref{tab:pilot_results}.

\begin{table}[h!]
    \centering
    \footnotesize
    \caption{\textbf{Pilot Study Case-Specific Rubric Results.} Summary of automated evaluation performance metrics across the clinical scenarios. Results compare AMIE and baseline PCPs across six experimental arms (text-only, audio-only, and video+audio), showing mean scores and 95\% bootstrap confidence intervals.}
    \label{tab:pilot_results}
    \renewcommand{\arraystretch}{1.5}
    \begin{tabular*}{\textwidth}{@{\extracolsep{\fill}} @{} >{\raggedright\arraybackslash}m{0.22\textwidth} c c c c c c @{}}
        \toprule
        & \multicolumn{3}{c}{\textbf{AMIE}} & \multicolumn{3}{c}{\textbf{Baseline PCPs}} \\
        \cmidrule(lr){2-4} \cmidrule(lr){5-7}
        \textbf{Evaluation Metric} & \textbf{Text} & \textbf{Audio} & \textbf{Video+Audio} & \textbf{Text} & \textbf{Audio} & \textbf{Video+Audio} \\
        \midrule
        \textbf{History Taking} & $\vcenter{\hbox{\shortstack{78.3\% \\ \tiny{[71.7\%, 85.8\%]}}}}$ & $\vcenter{\hbox{\shortstack{78.3\% \\ \tiny{[68.3\%, 87.5\%]}}}}$ & $\vcenter{\hbox{\shortstack{\textbf{81.7\%} \\ \tiny{[70.8\%, 90.8\%]}}}}$ & $\vcenter{\hbox{\shortstack{57.5\% \\ \tiny{[49.2\%, 67.5\%]}}}}$ & $\vcenter{\hbox{\shortstack{77.5\% \\ \tiny{[66.7\%, 87.5\%]}}}}$ & $\vcenter{\hbox{\shortstack{72.5\% \\ \tiny{[58.3\%, 85.0\%]}}}}$ \\
        \textbf{Perception and Examination} & $\vcenter{\hbox{\shortstack{42.5\% \\ \tiny{[27.5\%, 57.5\%]}}}}$ & $\vcenter{\hbox{\shortstack{52.1\% \\ \tiny{[38.3\%, 65.4\%]}}}}$ & $\vcenter{\hbox{\shortstack{\textbf{70.4\%} \\ \tiny{[59.2\%, 81.7\%]}}}}$ & $\vcenter{\hbox{\shortstack{26.2\% \\ \tiny{[13.8\%, 38.8\%]}}}}$ & $\vcenter{\hbox{\shortstack{37.9\% \\ \tiny{[25.0\%, 52.1\%]}}}}$ & $\vcenter{\hbox{\shortstack{55.0\% \\ \tiny{[38.8\%, 71.2\%]}}}}$ \\
        \textbf{Clinical Reasoning and Diagnosis} & $\vcenter{\hbox{\shortstack{92.5\% \\ \tiny{[86.2\%, 97.5\%]}}}}$ & $\vcenter{\hbox{\shortstack{90.0\% \\ \tiny{[81.2\%, 97.5\%]}}}}$ & $\vcenter{\hbox{\shortstack{\textbf{97.5\%} \\ \tiny{[93.8\%, 100.0\%]}}}}$ & $\vcenter{\hbox{\shortstack{75.0\% \\ \tiny{[61.2\%, 87.5\%]}}}}$ & $\vcenter{\hbox{\shortstack{80.0\% \\ \tiny{[68.8\%, 90.0\%]}}}}$ & $\vcenter{\hbox{\shortstack{76.2\% \\ \tiny{[61.3\%, 90.0\%]}}}}$ \\
        \textbf{Treatment Plan and Management} & $\vcenter{\hbox{\shortstack{61.3\% \\ \tiny{[51.2\%, 71.2\%]}}}}$ & $\vcenter{\hbox{\shortstack{61.3\% \\ \tiny{[47.5\%, 73.8\%]}}}}$ & $\vcenter{\hbox{\shortstack{62.5\% \\ \tiny{[52.5\%, 72.5\%]}}}}$ & $\vcenter{\hbox{\shortstack{53.8\% \\ \tiny{[40.0\%, 67.5\%]}}}}$ & $\vcenter{\hbox{\shortstack{63.7\% \\ \tiny{[50.0\%, 75.0\%]}}}}$ & $\vcenter{\hbox{\shortstack{\textbf{67.5\%} \\ \tiny{[53.8\%, 81.2\%]}}}}$ \\
        \textbf{Patient Communication and Education} & $\vcenter{\hbox{\shortstack{\textbf{95.0\%} \\ \tiny{[87.5\%, 100.0\%]}}}}$ & $\vcenter{\hbox{\shortstack{\textbf{95.0\%} \\ \tiny{[87.5\%, 100.0\%]}}}}$ & $\vcenter{\hbox{\shortstack{90.0\% \\ \tiny{[82.5\%, 97.5\%]}}}}$ & $\vcenter{\hbox{\shortstack{72.5\% \\ \tiny{[62.5\%, 85.0\%]}}}}$ & $\vcenter{\hbox{\shortstack{70.0\% \\ \tiny{[60.0\%, 80.1\%]}}}}$ & $\vcenter{\hbox{\shortstack{72.5\% \\ \tiny{[52.5\%, 90.0\%]}}}}$ \\
        \midrule
        \textbf{Overall Rubric Score} & $\vcenter{\hbox{\shortstack{72.4\% \\ \tiny{[67.1\%, 77.1\%]}}}}$ & $\vcenter{\hbox{\shortstack{73.7\% \\ \tiny{[66.7\%, 79.7\%]}}}}$ & $\vcenter{\hbox{\shortstack{\textbf{79.7\%} \\ \tiny{[74.0\%, 85.2\%]}}}}$ & $\vcenter{\hbox{\shortstack{55.6\% \\ \tiny{[48.5\%, 62.5\%]}}}}$ & $\vcenter{\hbox{\shortstack{66.7\% \\ \tiny{[60.5\%, 73.0\%]}}}}$ & $\vcenter{\hbox{\shortstack{68.9\% \\ \tiny{[58.2\%, 79.0\%]}}}}$ \\
        \bottomrule
    \end{tabular*}
\end{table}

\clearpage
\section{OSCE Study Details}
\label{appendix:osce_study_details}

We conducted a randomized, partially blinded study comparing the performance of various configurations of AMIE against PCPs on diagnostic dialogue in a synchronous video-based telehealth setting. We adapted the virtual Objective Structured Clinical Examination (OSCE) framework for evaluating conversational AI introduced in \citet{tu2025towards}, designing 120 scenario packs. The scenario packs and evaluation rubrics were enriched to ensure each scenario included potential audio-visual cues and guided examination components.

An initial pilot of 20 scenarios was conducted across 6 arms (AMIE Video, AMIE Audio, AMIE Text, PCP Video, PCP Audio, PCP Text). This pilot served as a preliminary validation of the system, a preparatory run to familiarize patient actors and PCPs with the interface, and a smaller-scale ablation to isolate the benefits of audio alone from video (\cref{appendix:pilot_run}). For the full study run, 100 scenarios were conducted with 3 arms (AMIE in text-only and video+audio settings, and PCPs in the video+audio setting). All sessions were assessed by clinical evaluators.

\subsection{Remote Video-based OSCE Examination}
\label{sec:study_design}

The OSCE is a standardized practical assessment widely used in healthcare training to evaluate clinical skills and competencies by simulating real-world practice. Unlike traditional knowledge-based examinations, the OSCE assesses practical skills of clinical encounters, typically involving candidates rotating through a series of timed stations where they encounter a trained patient actor portraying a specific clinical scenario. Test takers perform designated tasks such as taking a medical history, conducting a physical examination, interpreting results, or counseling the patient. Examiners observe these interactions and score the test taker's performance against detailed, predefined checklists that assess the specific skill components in each scenario.

\begin{figure}[h!]
    \centering
    \includegraphics[width=\textwidth]{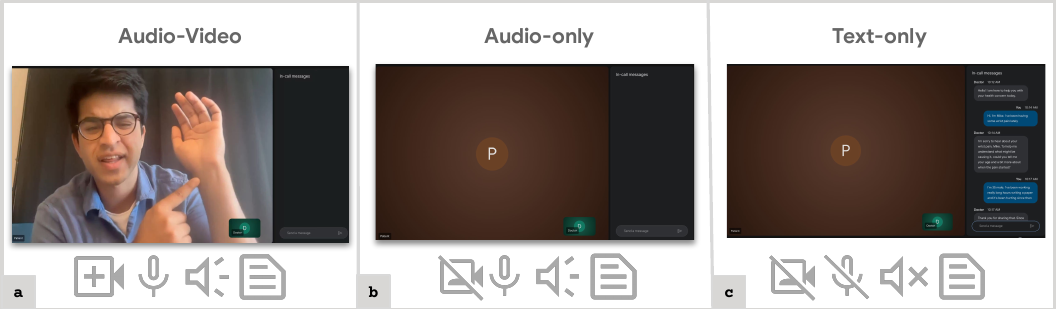}
    \vspace{0.1cm}
    \caption{\textbf{OSCE Study UI.} As seen by the PCP or AMIE. In all cases, the PCP and AMIE do not have their video on. a) In the Audio-Video arm, the patient actor can be seen and spoken to and heard through audio. b) In the audio-only arm (which was conducted only in the pilot phase) the patient actor cannot be seen, but communication still occurs over audio. c) In the text-only arm, the patient actor can be neither seen nor heard, and all communication occurs over the text chat window.}
    \label{figure:patient_actor_ui}
\end{figure}

Our study was heavily inspired by this OSCE framework, focusing on scenarios which could be performed in a virtual, video-based telehealth setting (\cref{sec:methods:scenario_packs}). Scenarios were prepared by external clinical providers and encompassed several broad clinical domains to ensure comprehensive coverage and diversity of audio-visual cues relevant to telehealth (with some exceptions based on actability). During the study, patient actors (\cref{sec:methods:patient_actors}) reviewed these scenario packs and used a video call interface similar to Google Meet (\cref{figure:patient_actor_ui}) to interact with AMIE or a PCP in a randomized sequence.

To assist with blinding, PCPs were asked to keep their cameras off; however, full blinding was not possible due to differing communication styles and voices between PCPs and AMIE. We note that the requirement for PCPs to have their cameras off may have impacted their ability to establish rapport and empathy compared to a typical audiovisual interaction, and this difference should be considered when interpreting findings related to communication and partnership building.

Following the conversation, the PCPs and AMIE each provided a post-questionnaire (\cref{box:post_questionnaire,sec:post_questionnaires}), which included their differential diagnosis and management recommendations. Patient actors rated the conversational quality of their interaction with either AMIE or the PCP. Finally, an independent panel of clinical evaluators (\cref{sec:methods:evaluators}) comprised of board-certified PCPs evaluated conversation quality and clinical accuracy from the full video, transcript, and post-questionnaires, using the review interface shown in \cref{figure:evaluator_ui}.

\subsection{Scenario Packs}
\label{sec:methods:scenario_packs}
This study utilized 120 distinct clinical case scenarios relevant to telehealth consultations for primary care conditions in the US. The scenarios were first developed by clinical experts from a Canadian organization with extensive experience in preparing scenario packs and simulated patients for OSCE examinations. Subsequently, each scenario underwent independent verification by one or more US board-certified PCPs.
Each scenario pack was associated with a ground truth diagnosis, a set of plausible alternative diagnoses, ground truth investigations and management, and a case-specific evaluation rubric (\cref{appendix:case_specific_rubric_example}). The scenario packs represented a range of conditions, focused on initial patient presentations, and were distributed across five domains where physical assessments are utilized and feasible in telehealth based on HHS Guidelines\footnote{https://telehealth.hhs.gov/providers/preparing-patients-for-telehealth/telehealth-physical-exam}: cardiopulmonary, abdominal, HEENT (head, eyes, ear, nose, throat), neurology and psychiatry, and musculoskeletal. Each domain included two ``non-disease'' cases for which the presenting complaint had resolved. Within each domain, scenarios and conditions were structured so that a broad breadth of physical cues and guided examination proficiencies could be assessed.
We intentionally excluded the following domains from the study: pediatrics, dermatology (limited ability for patient actors to demonstrate), routine chronic care follow-ups and annual physicals (limited ability to assess differential diagnosis), obstetrics, breast/gynecology/genital areas (sensitive body parts for patient actors), and inpatient scenarios.

\subsection{Patient Actors}
\label{sec:methods:patient_actors}

To ensure clinical realism and standardization, consultations were conducted by a dedicated cohort of 15 professional patient actors who reviewed their assigned clinical scenario packs in advance and executed standardized portrayals of patient affect, physical examination features, and guided examination maneuvers. Each clinical scenario was strictly matched to the actor's biological sex. 

The participating cohort comprised 7 females (46.7\%) and 8 males (53.3\%). The cohort spanned a diverse distribution of racial backgrounds and age groups (age range: 20--75 years; median age bracket: 30--40 years). Prior to the primary 100-scenario study, patient actors participated in a 20-scenario pilot evaluation (\cref{appendix:pilot_run}) to familiarize themselves with the Google Meet-style interface and calibrate their delivery across the study arms.

\subsection{Patient Actor Rubrics}
\label{sec:patient_actor_rubrics}

Following a consultation with the provider, patient actors provided answers to established criteria from GMCPQ (General Medical Council Patient Questionnaire), PACES (Practical Assessment of Clinical Examination Skills), and PCCBP (Patient-Centered Communication Best Practice) \citep{tu2025towards}. To elicit opinions on the different interaction modalities, the patient actors were additionally asked the following set of questions, providing a five-point Likert rating from ``Strongly Disagree'' to ``Strongly Agree'':
\begin{itemize}
    \item \emph{I was able to effectively communicate my health concerns and condition through this communication method.}
    \item \emph{I felt the clinician fully understood my concerns through this communication method.}
    \item \emph{I found this communication method easy and convenient to use.}
\end{itemize}

\subsection{Baseline Primary Care Physicians}
\label{sec:methods:pcps}

To establish a relevant human baseline for comparison, consultations across the PCP study arm were conducted by a cohort of 10 board-certified PCPs selected from a wider pool of potential PCPs based on type of board certification, type of medical qualification, and experience with urgent care or telemedicine care. 

The participating cohort comprised 10 physicians with diverse educational backgrounds: four held MDs from US medical schools, four held DOs from US osteopathic schools, and two held international MDs. Regarding board certification, eight were certified in Family Medicine and two in Internal Medicine. On average, the cohort had 4.5 years of post-residency clinical experience (range: 1–9 years). This distribution is broadly representative of US PCPs. Prior to initiating the primary 100-scenario study, participating PCPs were involved in a 20-scenario pilot evaluation (\cref{appendix:pilot_run}) to familiarize themselves with the protocol and Google Meet-style interface, mitigating technical issues during the main evaluation.

\subsection{Post Questionnaires}
\label{sec:post_questionnaires}

Following each simulated OSCE encounter, the respective clinical provider, whether PCP or an AMIE agent, completed a structured post-encounter clinical questionnaire documenting their differential diagnosis, clinical observations, and management planning (\cref{box:post_questionnaire}). Because this step is less latency-constrained than live conversations, the AMIE system utilizes inference-scaling via multiple drafts and synthesis (as in Lievin et al. \cite{lievin2026towards}) to improve the quality of its recommendations.

\subsection{Clinical Evaluators}
\label{sec:methods:evaluators}

All consultation transcripts, video recordings, and post-encounter questionnaires were reviewed by an external panel of 20 independent clinical evaluators. Unlike prior studies of the AMIE system \citep{tu2025towards,lievin2026towards} that utilized specialist physicians for evaluation, this study used board-certified PCPs as the clinical raters. This provided a more appropriate set of peers to evaluate clinical encounters within the telemedicine primary care setting.

To distinguish this evaluator panel from the baseline PCPs who participated in the OSCE encounters (\cref{sec:methods:pcps}), the clinical raters were recruited through a selective vetting process, emphasizing board certification in relevant specialties and extensive clinical experience including telemedicine, urgent care, or primary care. The evaluator cohort was entirely separate from the consultation cohort. The characteristics of the evaluators were as follows: type of medical degree (MD from US medical school 9, DO from US osteopathic school 6, MD from international medical school 5), board certification (Family Medicine 13, Internal Medicine 5, Emergency Medicine 2), and mean years since residency 14.95 years (range 8 to 25). 

\begin{figure}[h!]
    \centering
    \includegraphics[width=\textwidth]{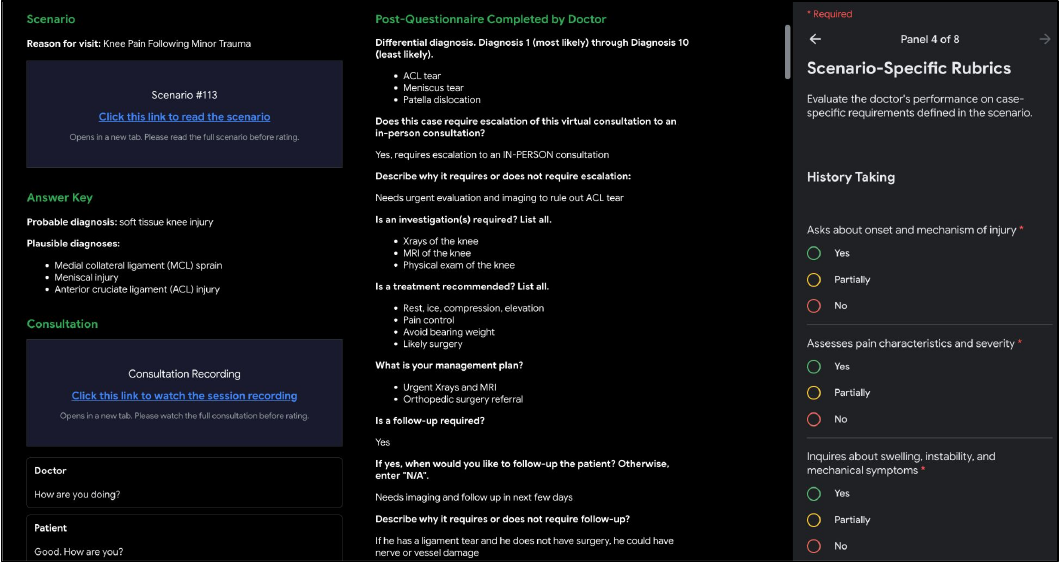}
    \vspace{0.1cm}
    \caption{\textbf{Clinical Evaluator Interface.} The web-based evaluation interface utilized by the independent panel of expert primary care evaluators. Raters are provided with the full scenario, consultation video recording, corresponding text transcript, and provider's completed post-encounter clinical questionnaire. The right-hand panel displays the structured evaluation rubrics, including overarching general competency scales (DDx Accuracy, Diagnosis \& Management, PACES, PCCBP, Audio-Visual Examination) and the 20-point case-specific rubrics.}
    \label{figure:evaluator_ui}
\end{figure}

Evaluators provided ratings on both general criteria (\cref{sec:general_rubrics}) and case-specific criteria (\cref{sec:case_specific_rubrics}) using the review interface shown in \cref{figure:evaluator_ui}. The same evaluator rated the performance of all study arms for a given scenario in a randomized order, and were not provided any explicit information about the source of each consultation.

\subsection{General Clinical Rating Rubrics}
\label{sec:general_rubrics}

Clinical evaluators assessed each encounter using established clinical communication and diagnostic reasoning rubrics adapted from prior benchmark evaluations \citep{tu2025towards, lievin2026towards}, primarily utilizing Likert scales across the general criteria. These overarching instruments evaluate multiple facets of clinical competence:
\begin{itemize}
    \item \textbf{Differential Diagnosis (DDx) Accuracy:} Assesses the ranking and inclusion of ground-truth clinical conditions \citep{tu2025towards}.
    \item \textbf{Diagnosis \& Management Rubric:} Measures the appropriateness and comprehensiveness of proposed diagnoses, investigations, treatment recommendations, escalation decisions, and follow-up planning \citep{tu2025towards,lievin2026towards}.
    \item \textbf{Patient-Centered Communication Scales (PACES):} Evaluates clinical history-taking skills, empathy, and the effectiveness of explaining information \citep{tu2025towards}.
    \item \textbf{Patient-Centered Care Benchmarking Protocol (PCCBP):} Gauges the provider's alignment with shared decision-making and patient education principles \citep{tu2025towards}.
\end{itemize}

To supplement these general instruments for synchronous video encounters, we incorporated a targeted set of telehealth evaluation criteria adapted from Sartori et al. \cite{sartori2020telehealth}. These supplementary items evaluate the provider's perceptual acuity in utilizing live video streams for information gathering and their ability to proactively partner with patients to execute guided physical examination maneuvers (\cref{appendix:av_telehealth_rubric}).

\subsection{Case-Specific Clinical Rating Rubrics}
\label{sec:case_specific_rubrics}

To evaluate clinical performance with sufficient granularity and nuance, each scenario was paired with a 20-point case-specific rubric, as in \citet{shah2026towards}. An example rubric is shown in \cref{appendix:case_specific_rubric_example}. During the grading process, evaluators could assess each individual criterion as fully met, partially met, or not met, corresponding to the award of full (1.0 pt), half (0.5 pt), or zero (0.0 pt) points, respectively.

The criteria for each case were standardized across five clinical domains:
\begin{itemize}
    \item \textbf{History Taking:} Evaluates the thoroughness and structure of the clinical interview, including eliciting key symptoms and their characteristics, eliciting risk factors, reviewing past medical history, and identifying potential red flags or warning signs of more serious or acute conditions.
    \item \textbf{Perception and Examination:} Assesses perceptual acuity regarding non-verbal cues, general patient affect, physical distress manifestations, and virtual examination maneuvers.
    \item \textbf{Clinical Reasoning and Diagnosis:} Evaluates diagnostic accuracy and the capacity to synthesize appropriate differential diagnoses spanning probable, plausible, and ``must-not-miss'' conditions.
    \item \textbf{Treatment Plan and Management:} Evaluates the clinical appropriateness, safety, and urgency of recommendations, including immediate management steps, diagnostic investigations, and follow-up planning.
    \item \textbf{Patient Communication and Education:} Assesses interpersonal communication quality, including the effectiveness of conveying clinical seriousness, and provides appropriate reassurance, establishing clear, actionable next steps.
\end{itemize}
\clearpage
\section{Scenario and Ground Truth Example}
\label{appendix:case_specific_rubric_example}

\begin{table}[h!]
\centering
\caption{\textbf{Example Scenario Details.} Complete scenario context details for a gastritis scenario (the corresponding clinical ground truth and evaluation rubric are shown in \cref{tab:ground_truth_rubric_2}).}
\label{tab:scenario_details_2}
\scriptsize
\renewcommand{\arraystretch}{1.1}
\setlength{\tabcolsep}{3pt}
\begin{tabular}{p{0.18\textwidth}p{0.80\textwidth}}
\toprule
\raggedright\textbf{Patient} & Maria Lopez (38-year-old Female, Hispanic, New York) \\
\raggedright\textbf{Medical Specialty} & Abdominal \\
\raggedright\textbf{Presenting Complaint} & You have been experiencing upper abdominal discomfort and nausea for the past week. \\
\raggedright\textbf{Initial Presentation} & "Hi doctor, I’ve had this dull ache in my upper stomach area for about a week now. It’s not terrible pain, but it’s constant, and I’ve been feeling nauseous too. Sometimes I burp a lot, and the discomfort seems worse after I eat. It’s starting to worry me." \\
\midrule
\raggedright\textbf{Info to be Given if Asked} & \textbf{Onset and Duration:} The discomfort started 7 days ago. Initially, it was mild, but it has become more noticeable and persistent over the past 3–4 days. \newline \textbf{Location and Character of Pain:} Pain is located in the upper central abdomen (epigastric region). \textbullet{} Describes it as a dull, burning ache rather than sharp pain. \newline \textbf{Aggravating and Relieving Factors:} Pain worsens after meals, especially spicy foods or coffee. \textbullet{} Slight relief when resting or taking antacids. \newline \textbf{Associated Symptoms:} Nausea, especially in the morning. \textbullet{} Occasional bloating and frequent burping. \textbullet{} No vomiting or blood in vomit \textbullet{} No black, tarry stools or visible blood in the stool. \textbullet{} No trouble swallowing. \textbullet{} No recent weight loss or appetite loss. \textbullet{} No cough or respiratory symptoms. \textbullet{} No chest pain or palpitations. \newline \textbf{Dietary Triggers:} Drinks 2–3 cups of coffee daily. \textbullet{} Recently started eating more takeout (pizza, fried foods) due to a busy schedule. \newline \textbf{Medication Use:} Has been taking ibuprofen almost daily for the past 2 weeks for knee pain. \textbullet{} Occasional use of over-the-counter antacids with mild relief. \newline \textbf{Red Flag Symptoms (Absent):} No severe, sudden-onset abdominal pain. \textbullet{} No vomiting of blood or coffee-ground material. \textbullet{} No unexplained weight loss. \textbullet{} No fever or chills. \\
\midrule
\raggedright\textbf{Clinical History} & \textbf{Past Medical History:} Mild knee osteoarthritis. \textbullet{} No known ulcers or previous gastrointestinal disease. \newline \textbf{Current Medications:} Ibuprofen 400 mg daily (taken for the past 2 weeks). \textbullet{} Antacid tablets as needed. \textbullet{} Daily multivitamin. \newline \textbf{Allergies \& Adverse Reactions:} No known drug allergies. \newline \textbf{Relevant Previous Medications:} Ibuprofen 400 mg daily (taken for the past 2 weeks). \textbullet{} Antacid tablets as needed. \textbullet{} Daily multivitamin. \\
\midrule
\raggedright\textbf{Social \& Family History} & \textbf{Social \& Personal Circumstances:} Works as an office administrator. \textbullet{} Non-smoker. \textbullet{} Drinks wine socially (1–2 glasses on weekends). \newline \textbf{Family History:} No known family history regarding symptoms. \newline \textbf{Occupational History:} Works as an office administrator. \newline \textbf{Lifestyle:} Works as an office administrator. \textbullet{} Non-smoker. \textbullet{} Drinks wine socially (1–2 glasses on weekends). \newline \textbf{Travel History:} No recent travel. \\
\midrule
\raggedright\textbf{Concerns \& Questions} & \textbf{Concerns:} “Could this be something serious, like an ulcer or even cancer?” \newline \textbf{Expectations:} “I’d like to know if I need tests or medications to help with the pain.” \newline \textbf{Wishes:} “I just want this discomfort to stop so I can eat normally again.” \newline \textbf{Specific Questions:} Is it safe to keep taking ibuprofen for my knee pain? \textbullet{} Could this be caused by stress or something I ate? \textbullet{} Do I need an endoscopy or any other tests? \\
\midrule
\raggedright\textbf{Patient Instructions} & \textbf{1. General Instruction:} Speak in a calm, slightly tired tone, occasionally holding your upper abdomen when describing discomfort. Avoid dramatic expressions of pain — keep it mild to moderate. \newline \textbf{2. Abdominal Pain Demonstration (If Asked):} Point to the area just below the breastbone and slightly above the belly button. \textbullet{} Say: “The ache is mostly here, right in the middle.” \newline \textbf{3. Aggravation with Food (If Asked):} Rub your stomach gently and say: “It feels worse about 30 minutes after I eat, especially if it’s spicy or fried food.” \newline \textbf{4. Medication History (If Asked):} Look slightly concerned and say: “I’ve been taking ibuprofen every day for the past few weeks because of my knee pain. Could that be making it worse?” \newline \textbf{5. Nausea Description (If Asked):} Make a mild grimace and say: “I feel queasy most mornings, but I haven’t actually vomited.” \\
\bottomrule
\end{tabular}
\end{table}

\begin{table}[h!]
\centering
\caption{\textbf{Example Ground Truth \& Evaluation Rubric.} Clinical ground truth and case-specific evaluation rubric for a gastritis scenario (the corresponding scenario details are shown in \cref{tab:scenario_details_2}).}
\label{tab:ground_truth_rubric_2}
\scriptsize
\renewcommand{\arraystretch}{1.1}
\setlength{\tabcolsep}{3pt}
\begin{tabular}{p{0.18\textwidth}p{0.77\textwidth}r}
\toprule
\raggedright\textbf{Probable Diagnosis} & \multicolumn{2}{p{0.79\textwidth}}{Gastritis} \\
\raggedright\textbf{Plausible Alternative Diagnoses} & \multicolumn{2}{p{0.79\textwidth}}{Peptic Ulcer Disease, Functional Dyspepsia} \\
\midrule
\raggedright\textbf{Investigations \& Action Plan} & \multicolumn{2}{p{0.79\textwidth}}{No urgent tests needed unless red flags are present. \textbullet{} Consider CBC to check for anemia. \textbullet{} Consider H. pylori breath test or stool antigen test. \textbullet{} Consider endoscopy - only if symptoms are severe or persistent, or there are warning signs.} \\
\midrule
\raggedright\textbf{Management} & \multicolumn{2}{p{0.79\textwidth}}{\textbf{Medical Treatments:} Stop/minimize NSAID use - consider acetaminophen for knee pain instead. Start proton pump inhibitor (PPI) for suspected gastritis. \newline \textbf{Lifestyle Modifications:} Avoid caffeine, alcohol, spicy foods, and greasy meals. Eat smaller, more frequent meals. Remain upright after meals, Avoid eating right before bed. Reduce stress where possible. \newline \textbf{Follow-up:} Reassess symptoms in 2–4 weeks. If worsening or red flags develop, proceed to urgent investigation.} \\
\midrule
\multicolumn{3}{l}{\textbf{Case-Specific Evaluation Rubric}} \\
\midrule
\multicolumn{3}{l}{\textbf{History Taking}} \\
& Asks about onset, duration, and location of discomfort. & 1 \\
& Inquires about aggravating/relieving factors and meal relationship. & 1 \\
& Asks about associated symptoms (nausea, vomiting, bleeding). & 1 \\
& Reviews medication history, especially NSAIDs. & 1 \\
& Screens for red flags (weight loss, severe pain, anemia). & 1 \\
& Reviews past medical and family history of ulcers or GI disease. & 1 \\
\midrule
\multicolumn{3}{l}{\textbf{Perception and Examination}} \\
& Observes pointing to epigastric region correctly. & 1 \\
& Assesses visible discomfort when discussing symptoms. & 1 \\
& Checks for systemic signs (fatigue, pallor). & 1 \\
& Notes absence of severe distress or alarming symptoms. & 1 \\
\midrule
\multicolumn{3}{l}{\textbf{Clinical Reasoning and Diagnosis}} \\
& Recognizes likely gastritis, possibly NSAID-induced. & 2 \\
& Appropriately rules out urgent conditions (ulcer, bleeding, malignancy). & 2 \\
\midrule
\multicolumn{3}{l}{\textbf{Treatment Plan and Management}} \\
& Stops or minimizes NSAID use and switches to safer alternative. & 1 \\
& Starts PPI for suspected gastritis. & 1 \\
& Advises on dietary modifications and trigger avoidance. & 1 \\
& Plans appropriate follow-up timeline and tests if needed. & 1 \\
\midrule
\multicolumn{3}{l}{\textbf{Patient Communication and Education}} \\
& Explains probable cause (NSAIDs, diet) clearly to patient. & 1 \\
& Provides reassurance and clear home care instructions. & 1 \\
\bottomrule
\end{tabular}
\end{table}
\clearpage
\section{Telehealth Audio-Visual Examination Rubric}
\label{appendix:av_telehealth_rubric}

To evaluate the unique clinical competencies required during synchronous video encounters, we supplemented the established general evaluation rubrics with targeted telehealth criteria adapted from \citet{sartori2020telehealth}. These items specifically assess the provider's perceptual acuity in utilizing live video streams for information gathering, their ability to proactively partner with patients to execute guided physical examination maneuvers, and provide open-ended qualitative feedback regarding interface utilization. The complete evaluation items, rating scales, and adaptation rationale are detailed in \cref{tab:av_telehealth_rubric}.

\vspace{1em}

\begin{table}[h!]
    \centering
    \footnotesize
    \caption{\textbf{Supplementary Telehealth OSCE Evaluation Rubric.} Evaluation items, scoring scales, and rationale incorporated into the expert rater general rubrics to assess audio-visual symptom observation and patient-assisted physical examinations. Criteria 1 and 2 are adapted from Sartori et al. \cite{sartori2020telehealth}, while Criterion 3 was added to capture open-ended qualitative feedback.}
    \label{tab:av_telehealth_rubric}
    \begin{tabular}{@{} >{\raggedright\arraybackslash}p{0.55\textwidth} >{\raggedright\arraybackslash}p{0.18\textwidth} >{\raggedright\arraybackslash}p{0.22\textwidth} @{}}
        \toprule
        \textbf{Checklist Item \& Rater Guidance} & \textbf{Rating Scale} & \textbf{Source} \\
        \midrule
        \textbf{To what extent did the doctor utilize live video to augment information gathering?} \par\vspace{0.4em} \emph{When rating this item, please consider whether the doctor perceived all the relevant visual and audio cues from the live video, for example if they identified reproducible symptoms or visually assessed patient non-verbal cues.} & 5-point scale \par\vspace{0.4em} (Very Poor, Poor, Fair, Good, Excellent) & \textbf{Adapt:} Adapted from \citet{sartori2020telehealth} to better fit the technical constraints of this study. \\
        \midrule
        \textbf{To what extent did the doctor partner with the patient to perform physical examination?} \par\vspace{0.4em} \emph{When rating this item, please consider whether the doctor asked the patient to perform maneuvers or access peripheral monitoring device (home blood pressure cuff, wearable device, glucometer), followed by verbal confirmation of findings.} & 5-point scale \par\vspace{0.4em} (Very Poor, Poor, Fair, Good, Excellent) & \textbf{Keep:} Kept directly from \citet{sartori2020telehealth} source framework. \\
        \midrule
        \textbf{What did the agent do well or not do well in the video interface (for example, unexpected, missed, etc.)?} & Open-ended free text & \textbf{Added:} Incorporated for qualitative interpretation of video interface utilization. \\
        \bottomrule
    \end{tabular}
\end{table}
\clearpage
\section{Post-Encounter Clinical Questionnaire}
\label{appendix:post_questionnaire}

\begin{mdframed}[backgroundcolor=black!5!white, linecolor=black!60!white, linewidth=1pt, roundcorner=1mm]
\footnotesize
\textbf{\large Post-Encounter Clinical Questionnaire Structure}
\label{box:post_questionnaire}

\noindent\rule{\textwidth}{0.5pt}

\noindent\textbf{\normalsize 1. Differential Diagnosis}

\vspace{0.2em}
\noindent\begin{tabular}{@{}p{0.48\textwidth} p{0.48\textwidth}@{}}
    \textbullet~\textbf{Diagnosis 1 (most likely) *} & \textbullet~\textbf{Diagnosis 6} \\
    \textbullet~\textbf{Diagnosis 2 *} & \textbullet~\textbf{Diagnosis 7} \\
    \textbullet~\textbf{Diagnosis 3 *} & \textbullet~\textbf{Diagnosis 8} \\
    \textbullet~\textbf{Diagnosis 4} & \textbullet~\textbf{Diagnosis 9} \\
    \textbullet~\textbf{Diagnosis 5} & \textbullet~\textbf{Diagnosis 10 (least likely)} \\
\end{tabular}

\vspace{0.4em}
\noindent\textbf{\normalsize 2. Observation}

\begin{itemize}[leftmargin=*, noitemsep, topsep=0pt, parsep=0pt, partopsep=0pt]
    \item \textbf{Audio-Visual Cues:} If this was an audio or video interaction, please describe any audio-visual cues you noticed when interacting with the patient and how they informed your reasoning about this case? Enter N/A if you did not notice any particular cues or if this was a text interaction.
    \item \textbf{Desired Observations:} If this was a text or audio interaction, please describe what you would have liked to observe to help you reason about the case. Enter N/A if this does not apply.
\end{itemize}

\vspace{0.4em}
\noindent\textbf{\normalsize 3. Clinical Management}

\begin{itemize}[leftmargin=*, noitemsep, topsep=0pt, parsep=0pt, partopsep=0pt]
    \item \textbf{In-Person Escalation:} Does this case require escalation of this virtual consultation to an in-person consultation? \\
    $\square$ No, does not require escalation \qquad $\square$ Yes, requires escalation to an IN-PERSON consultation
    \item \textbf{Escalation Reasoning:} Describe why it requires or does not require escalation:
    
    \item \textbf{Required Investigations:} Is an investigation(s) required? List all. \\
    \emph{This could include imaging, lab tests, etc. It should also include any other examinations, tests, or investigations you believe to be necessary but were unable to perform due to the virtual nature of the consultation. Format your response as a simple bullet list starting with dashes. (e.g., - Some test or investigation.)}

    \item \textbf{Recommended Treatments:} Is a treatment recommended? List all.
    \item \textbf{Management Plan:} What is your management plan?
    
    \item \textbf{Alternative Management Plans:} Is there more than one acceptable management plan for this case?\\ \quad $\square$ Yes \qquad $\square$ No
    \item \textbf{Alternative Management Reasoning:} If other management plans were acceptable, please describe the alternatives and explain your choice. If no other options were viable, please briefly explain why.
    
    \item \textbf{High-Level Trade-off Reasoning:} Briefly describe your qualitative, high-level reasoning around the cost, effectiveness and side effects for the management plan you are proposing. Try to highlight why you are recommending specific things to keep the cost manageable, side effects minimal or increase effectiveness of the treatment, and any trade-offs involved.
    
    \item \textbf{Prognosis Assessment:} Describe the prognosis with and without treatment of this patient.
    
    \item \textbf{Follow-up Requirement:} Is a follow-up required?\\ \quad $\square$ Yes \qquad $\square$ No \qquad $\square$ Unsure
    \item \textbf{Follow-up Timing:} If yes, when would you like to follow-up the patient? Otherwise, enter ``N/A''.
    \item \textbf{Follow-up Reasoning:} Describe why it requires or does not require follow-up?
    
    \item \textbf{Management Script:} Please describe the management script you considered for this case? In other words, the step by step reasoning you used in this SPECIFIC case which could be applied to another case with this same chief complaint?
    
    \item \textbf{Additional Remarks:} Is there anything else you would like to tell us? \emph{(Optional)}
\end{itemize}
\end{mdframed}
\clearpage
\section{User Interfaces}
\label{appendix_section:patient_actor_ui}

\clearpage
\section{Tabular OSCE Study Results}
\label{appendix:tabular_results}

\begin{table}[h!]
    \centering
    \scriptsize
    \caption{\textbf{OSCE Study Case-Specific Rubric Results.} Expert-evaluated performance of AMIE (Video), AMIE (Text) and PCP (Video). A two-sided Wilcoxon signed-rank test with Benjamini--Hochberg false discovery rate (FDR) correction is used to compare AMIE (Video) to the other arms. 95\% confidence intervals are computed via scenario-level bootstrapping (N=1000).}
    \label{tab:osce_case_rubric_results}
    \renewcommand{\arraystretch}{1.08}
    \begin{tabular*}{\textwidth}{@{\extracolsep{\fill}} @{} >{\raggedright\arraybackslash}m{0.24\textwidth} c c c c c c @{}}
        \toprule
        & & \multicolumn{3}{c}{\textbf{Performance [95\% CI]}} & \multicolumn{2}{c}{\textbf{FDR-adjusted $p$-value}} \\
        \cmidrule(lr){3-5} \cmidrule(lr){6-7}
        \textbf{Evaluation Domain} & \textbf{Scale} & \textbf{AMIE (Video)} & \textbf{AMIE (Text)} & \textbf{PCP (Video)} & \textbf{vs.~PCP} & \textbf{vs.~Text} \\
        \midrule
        \textbf{History Taking} & Rubric & \textbf{0.86} \tiny{[0.83, 0.89]} & 0.85 \tiny{[0.82, 0.88]} & 0.78 \tiny{[0.74, 0.82]} & $1.20 \times 10^{-3}$ & $0.098$ \\
        \textbf{Perception and Examination} & Rubric & \textbf{0.74} \tiny{[0.69, 0.80]} & 0.41 \tiny{[0.35, 0.48]} & 0.47 \tiny{[0.40, 0.54]} & $1.73 \times 10^{-9}$ & $3.32 \times 10^{-11}$ \\
        \textbf{Clinical Reasoning and Diagnosis} & Rubric & \textbf{0.90} \tiny{[0.87, 0.93]} & 0.87 \tiny{[0.83, 0.91]} & 0.76 \tiny{[0.70, 0.81]} & $2.00 \times 10^{-5}$ & $0.186$ \\
        \textbf{Treatment Plan and Management} & Rubric & \textbf{0.79} \tiny{[0.74, 0.84]} & 0.78 \tiny{[0.73, 0.82]} & 0.67 \tiny{[0.61, 0.71]} & $8.90 \times 10^{-5}$ & $0.331$ \\
        \textbf{Patient Communication and Education} & Rubric & \textbf{0.85} \tiny{[0.80, 0.90]} & 0.80 \tiny{[0.74, 0.85]} & 0.68 \tiny{[0.62, 0.74]} & $6.56 \times 10^{-7}$ & $0.114$ \\
        \midrule
        \textbf{Overall Rubric Score} & Rubric & \textbf{0.83} \tiny{[0.81, 0.86]} & 0.75 \tiny{[0.72, 0.78]} & 0.68 \tiny{[0.64, 0.72]} & $1.32 \times 10^{-9}$ & $1.19 \times 10^{-8}$ \\
        \bottomrule
    \end{tabular*}
\end{table}

\begin{table}[h!]
    \centering
    \scriptsize
    \caption{\textbf{OSCE General Expert Evaluation Results.} Expert-evaluated performance of AMIE (Video), AMIE (Text) and PCP (Video). For display and analysis, ratings are mapped to a 0--1 scale (mapping 5-point Likert ratings via $(S-1)/4$ and binary outcomes to $1$ for favorable and $0$ for non-favorable). For statistical comparison, a two-sided Wilcoxon signed-rank test with Benjamini--Hochberg false discovery rate (FDR) correction was used to compare AMIE (Video) to the other arms. Cases receiving `N/A' ratings on any arm were excluded from paired analysis. 95\% confidence intervals are computed via scenario-level bootstrapping (N=1000).}
    \label{tab:osce_general_expert_results}
    \renewcommand{\arraystretch}{1.08}
    \begin{tabular*}{\textwidth}{@{\extracolsep{\fill}} @{} >{\raggedright\arraybackslash}m{0.24\textwidth} c c c c c c @{}}
        \toprule
        & & \multicolumn{3}{c}{\textbf{Performance [95\% CI]}} & \multicolumn{2}{c}{\textbf{FDR-adjusted $p$-value}} \\
        \cmidrule(lr){3-5} \cmidrule(lr){6-7}
        \textbf{Evaluation Criteria} & \textbf{Scale} & \textbf{AMIE (Video)} & \textbf{AMIE (Text)} & \textbf{PCP (Video)} & \textbf{vs.~PCP} & \textbf{vs.~Text} \\
        \midrule
        \multicolumn{7}{@{}l}{\textbf{\textsf{Diagnostic Accuracy (DDx)}}} \\
        \quad Top-1 DDx accuracy & Binary & \textbf{0.91} \tiny{[0.85, 0.96]} & 0.90 \tiny{[0.84, 0.95]} & 0.77 \tiny{[0.68, 0.85]} & $0.039$ & $0.862$ \\
        \quad Top-3 DDx accuracy & Binary & \textbf{0.98} \tiny{[0.95, 1.00]} & 0.97 \tiny{[0.94, 1.00]} & 0.90 \tiny{[0.83, 0.96]} & $0.179$ & $0.862$ \\
        \quad DDx appropriateness & 5-point Likert & 0.89 \tiny{[0.85, 0.92]} & \textbf{0.89} \tiny{[0.86, 0.92]} & 0.80 \tiny{[0.75, 0.84]} & $1.05 \times 10^{-3}$ & $0.862$ \\
        \quad DDx comprehensiveness & 5-point Likert & \textbf{0.88} \tiny{[0.85, 0.92]} & 0.86 \tiny{[0.83, 0.90]} & 0.73 \tiny{[0.69, 0.78]} & $2.38 \times 10^{-6}$ & $0.494$ \\
        \addlinespace[0.25em]
        \multicolumn{7}{@{}l}{\textbf{\textsf{Management Plan \& Safety}}} \\
        \quad Management plan appropriateness & 5-point Likert & \textbf{0.79} \tiny{[0.74, 0.84]} & 0.77 \tiny{[0.70, 0.82]} & 0.68 \tiny{[0.62, 0.74]} & $7.55 \times 10^{-3}$ & $0.625$ \\
        \quad Investigation appropriateness & Binary & \textbf{0.93} \tiny{[0.88, 0.98]} & 0.89 \tiny{[0.82, 0.95]} & 0.82 \tiny{[0.74, 0.89]} & $0.089$ & $0.625$ \\
        \quad Avoid inappropriate investigations & Binary & 0.88 \tiny{[0.82, 0.94]} & \textbf{0.89} \tiny{[0.82, 0.95]} & 0.79 \tiny{[0.70, 0.87]} & $0.149$ & $0.862$ \\
        \quad Treatment appropriateness & Binary & \textbf{0.87} \tiny{[0.80, 0.93]} & 0.82 \tiny{[0.74, 0.89]} & 0.76 \tiny{[0.68, 0.84]} & $0.089$ & $0.590$ \\
        \quad Avoid inappropriate treatments & Binary & \textbf{0.90} \tiny{[0.84, 0.95]} & 0.86 \tiny{[0.79, 0.93]} & 0.82 \tiny{[0.74, 0.90]} & $0.189$ & $0.625$ \\
        \quad Escalation recommendation & Binary & \textbf{0.91} \tiny{[0.85, 0.96]} & 0.90 \tiny{[0.84, 0.96]} & 0.83 \tiny{[0.76, 0.91]} & $0.191$ & $0.862$ \\
        \quad Follow-up recommendation & Binary & \textbf{0.94} \tiny{[0.89, 0.98]} & 0.92 \tiny{[0.86, 0.97]} & 0.92 \tiny{[0.86, 0.97]} & $0.737$ & $0.862$ \\
        \quad Avoid confabulation & Binary & 0.93 \tiny{[0.88, 0.98]} & \textbf{0.95} \tiny{[0.91, 0.99]} & 0.93 \tiny{[0.88, 0.98]} & $1.0$ & $0.862$ \\
        \addlinespace[0.25em]
        \multicolumn{7}{@{}l}{\textbf{\textsf{Patient-Centered Communication (PCCBP)}}} \\
        \quad Foster relationship & 5-point Likert & \textbf{0.80} \tiny{[0.76, 0.83]} & 0.74 \tiny{[0.70, 0.78]} & 0.69 \tiny{[0.63, 0.74]} & $5.63 \times 10^{-3}$ & $0.088$ \\
        \quad Shared decision making & 5-point Likert & \textbf{0.78} \tiny{[0.74, 0.82]} & 0.74 \tiny{[0.70, 0.79]} & 0.67 \tiny{[0.62, 0.72]} & $8.49 \times 10^{-4}$ & $0.431$ \\
        \quad Patient teacher & 5-point Likert & \textbf{0.83} \tiny{[0.80, 0.86]} & 0.78 \tiny{[0.74, 0.82]} & 0.70 \tiny{[0.65, 0.74]} & $6.51 \times 10^{-5}$ & $0.182$ \\
        \quad Empathy (PCCBP) & 5-point Likert & \textbf{0.76} \tiny{[0.73, 0.80]} & 0.71 \tiny{[0.68, 0.75]} & 0.64 \tiny{[0.58, 0.70]} & $1.85 \times 10^{-3}$ & $0.266$ \\
        \quad Agenda setting & 5-point Likert & \textbf{0.77} \tiny{[0.72, 0.81]} & 0.70 \tiny{[0.66, 0.74]} & 0.67 \tiny{[0.62, 0.72]} & $5.16 \times 10^{-3}$ & $0.037$ \\
        \quad Active listening & 5-point Likert & \textbf{0.82} \tiny{[0.79, 0.85]} & 0.75 \tiny{[0.71, 0.79]} & 0.68 \tiny{[0.62, 0.73]} & $1.37 \times 10^{-4}$ & $0.071$ \\
        \addlinespace[0.25em]
        \multicolumn{7}{@{}l}{\textbf{\textsf{Clinical \& Communication Competencies (PACES)}}} \\
        \quad Eliciting presenting complaint & 5-point Likert & \textbf{0.95} \tiny{[0.92, 0.97]} & 0.92 \tiny{[0.90, 0.95]} & 0.82 \tiny{[0.78, 0.87]} & $4.83 \times 10^{-5}$ & $0.431$ \\
        \quad Eliciting systems review & 5-point Likert & \textbf{0.88} \tiny{[0.84, 0.91]} & 0.86 \tiny{[0.81, 0.89]} & 0.74 \tiny{[0.68, 0.80]} & $1.05 \times 10^{-3}$ & $0.494$ \\
        \quad Eliciting past medical history & 5-point Likert & 0.89 \tiny{[0.85, 0.92]} & \textbf{0.91} \tiny{[0.87, 0.94]} & 0.85 \tiny{[0.80, 0.89]} & $0.359$ & $0.526$ \\
        \quad Eliciting family history & 5-point Likert & \textbf{0.88} \tiny{[0.84, 0.93]} & 0.86 \tiny{[0.81, 0.91]} & 0.78 \tiny{[0.71, 0.85]} & $7.44 \times 10^{-3}$ & $0.494$ \\
        \quad Eliciting medication history & 5-point Likert & \textbf{0.90} \tiny{[0.87, 0.93]} & 0.88 \tiny{[0.84, 0.92]} & 0.85 \tiny{[0.79, 0.89]} & $0.149$ & $0.526$ \\
        \quad Explaining accurately & 5-point Likert & \textbf{0.91} \tiny{[0.89, 0.94]} & 0.91 \tiny{[0.88, 0.94]} & 0.81 \tiny{[0.77, 0.85]} & $1.25 \times 10^{-4}$ & $0.862$ \\
        \quad Explaining clearly & 5-point Likert & \textbf{0.94} \tiny{[0.91, 0.96]} & 0.90 \tiny{[0.87, 0.93]} & 0.77 \tiny{[0.72, 0.81]} & $2.38 \times 10^{-6}$ & $0.266$ \\
        \quad Structure \& time management & 5-point Likert & \textbf{0.91} \tiny{[0.88, 0.94]} & 0.89 \tiny{[0.86, 0.92]} & 0.73 \tiny{[0.67, 0.78]} & $2.49 \times 10^{-6}$ & $0.494$ \\
        \quad Explaining comprehensively & 5-point Likert & \textbf{0.87} \tiny{[0.84, 0.90]} & 0.83 \tiny{[0.79, 0.87]} & 0.71 \tiny{[0.66, 0.76]} & $1.15 \times 10^{-5}$ & $0.266$ \\
        \quad Interpersonal skills & 5-point Likert & \textbf{0.96} \tiny{[0.94, 0.97]} & 0.94 \tiny{[0.91, 0.96]} & 0.83 \tiny{[0.79, 0.87]} & $6.51 \times 10^{-5}$ & $0.526$ \\
        \quad Differential diagnosis (PACES) & 5-point Likert & \textbf{0.90} \tiny{[0.87, 0.93]} & 0.89 \tiny{[0.85, 0.93]} & 0.76 \tiny{[0.70, 0.81]} & $6.05 \times 10^{-4}$ & $0.862$ \\
        \quad Clinical judgement & 5-point Likert & \textbf{0.88} \tiny{[0.84, 0.92]} & 0.86 \tiny{[0.81, 0.90]} & 0.78 \tiny{[0.73, 0.83]} & $0.017$ & $0.494$ \\
        \quad Managing patient concerns & 5-point Likert & \textbf{0.88} \tiny{[0.85, 0.91]} & 0.81 \tiny{[0.76, 0.85]} & 0.79 \tiny{[0.74, 0.84]} & $4.36 \times 10^{-3}$ & $0.037$ \\
        \quad Confirming understanding & 5-point Likert & \textbf{0.80} \tiny{[0.76, 0.84]} & 0.71 \tiny{[0.66, 0.77]} & 0.69 \tiny{[0.63, 0.74]} & $6.05 \times 10^{-4}$ & $0.037$ \\
        \quad Showing empathy & 5-point Likert & \textbf{0.82} \tiny{[0.79, 0.85]} & 0.79 \tiny{[0.75, 0.82]} & 0.71 \tiny{[0.66, 0.76]} & $3.39 \times 10^{-3}$ & $0.305$ \\
        \quad Maintaining patient welfare & 5-point Likert & \textbf{0.91} \tiny{[0.88, 0.93]} & 0.87 \tiny{[0.83, 0.90]} & 0.81 \tiny{[0.76, 0.85]} & $2.14 \times 10^{-3}$ & $0.266$ \\
        \addlinespace[0.25em]
        \multicolumn{7}{@{}l}{\textbf{\textsf{Physical Observation \& Examination}}} \\
        \quad Physical observation & 5-point Likert & \textbf{0.77} \tiny{[0.70, 0.83]} & 0.30 \tiny{[0.21, 0.40]} & 0.51 \tiny{[0.42, 0.60]} & $1.37 \times 10^{-4}$ & $3.85 \times 10^{-6}$ \\
        \quad Guided examination & 5-point Likert & \textbf{0.72} \tiny{[0.67, 0.78]} & 0.51 \tiny{[0.44, 0.58]} & 0.39 \tiny{[0.32, 0.46]} & $2.46 \times 10^{-9}$ & $1.93 \times 10^{-5}$ \\
        \bottomrule
    \end{tabular*}
\end{table}

\begin{table}[h!]
    \centering
    \scriptsize
    \caption{\textbf{OSCE Patient Actor Ratings Results.} Patient actor evaluations of AMIE (Video), AMIE (Text) and PCP (Video). For display and analysis, ratings are mapped to a 0--1 scale (mapping 5-point Likert ratings via $(S-1)/4$ and binary outcomes to $1$ for favorable and $0$ for non-favorable). For statistical comparison, a two-sided Wilcoxon signed-rank test with Benjamini--Hochberg false discovery rate (FDR) correction was used to compare AMIE (Video) to the other arms. Cases receiving `N/A' ratings on any arm were excluded from paired analysis. 95\% confidence intervals are computed via scenario-level bootstrapping (N=1000).}
    \label{tab:osce_patient_actor_results}
    \renewcommand{\arraystretch}{1.08}
    \begin{tabular*}{\textwidth}{@{\extracolsep{\fill}} @{} >{\raggedright\arraybackslash}m{0.24\textwidth} c c c c c c @{}}
        \toprule
        & & \multicolumn{3}{c}{\textbf{Performance [95\% CI]}} & \multicolumn{2}{c}{\textbf{FDR-adjusted $p$-value}} \\
        \cmidrule(lr){3-5} \cmidrule(lr){6-7}
        \textbf{Patient Rating Criteria} & \textbf{Scale} & \textbf{AMIE (Video)} & \textbf{AMIE (Text)} & \textbf{PCP (Video)} & \textbf{vs.~PCP} & \textbf{vs.~Text} \\
        \midrule
        \multicolumn{7}{@{}l}{\textbf{\textsf{General Medical Council \& Trust (GMCPQ)}}} \\
        \quad Being polite & 5-point Likert & \textbf{0.89} \tiny{[0.85, 0.93]} & 0.85 \tiny{[0.82, 0.89]} & 0.88 \tiny{[0.83, 0.91]} & $0.775$ & $0.120$ \\
        \quad Making feel at ease & 5-point Likert & \textbf{0.84} \tiny{[0.80, 0.89]} & 0.76 \tiny{[0.72, 0.81]} & 0.82 \tiny{[0.77, 0.86]} & $0.750$ & $6.93 \times 10^{-4}$ \\
        \quad Listening to you & 5-point Likert & \textbf{0.87} \tiny{[0.82, 0.92]} & 0.83 \tiny{[0.79, 0.88]} & \textbf{0.87} \tiny{[0.82, 0.92]} & $0.952$ & $0.110$ \\
        \quad Assessing condition & 5-point Likert & \textbf{0.91} \tiny{[0.87, 0.94]} & 0.89 \tiny{[0.86, 0.93]} & 0.81 \tiny{[0.75, 0.86]} & $0.045$ & $0.321$ \\
        \quad Explaining condition & 5-point Likert & \textbf{0.88} \tiny{[0.84, 0.93]} & 0.84 \tiny{[0.80, 0.89]} & 0.79 \tiny{[0.73, 0.84]} & $0.045$ & $0.120$ \\
        \quad Involving in decisions & 5-point Likert & \textbf{0.79} \tiny{[0.74, 0.83]} & 0.78 \tiny{[0.74, 0.82]} & 0.76 \tiny{[0.70, 0.81]} & $0.750$ & $0.914$ \\
        \quad Providing treatment & 5-point Likert & 0.79 \tiny{[0.74, 0.84]} & 0.74 \tiny{[0.69, 0.78]} & \textbf{0.80} \tiny{[0.74, 0.84]} & $0.690$ & $0.128$ \\
        \quad Confidentiality & 5-point Likert & \textbf{0.87} \tiny{[0.83, 0.91]} & 0.84 \tiny{[0.80, 0.88]} & \textbf{0.87} \tiny{[0.83, 0.91]} & $0.750$ & $0.128$ \\
        \quad Honest \& trustworthy & 5-point Likert & 0.87 \tiny{[0.83, 0.91]} & 0.84 \tiny{[0.80, 0.88]} & \textbf{0.88} \tiny{[0.83, 0.92]} & $0.690$ & $0.128$ \\
        \quad Confident in care & Binary & \textbf{0.96} \tiny{[0.92, 1.00]} & 0.85 \tiny{[0.78, 0.91]} & 0.88 \tiny{[0.81, 0.94]} & $0.690$ & $0.128$ \\
        \quad Happy to see again & Binary & \textbf{0.91} \tiny{[0.85, 0.97]} & 0.74 \tiny{[0.65, 0.83]} & 0.81 \tiny{[0.73, 0.89]} & $0.618$ & $0.038$ \\
        \addlinespace[0.25em]
        \multicolumn{7}{@{}l}{\textbf{\textsf{Fostering Relationship (PCCBP)}}} \\
        \quad Build rapport & Binary & 0.81 \tiny{[0.73, 0.88]} & 0.66 \tiny{[0.57, 0.75]} & \textbf{0.82} \tiny{[0.74, 0.89]} & $0.830$ & $0.110$ \\
        \quad Open \& honest & Binary & \textbf{0.99} \tiny{[0.97, 1.00]} & 0.92 \tiny{[0.86, 0.97]} & 0.96 \tiny{[0.92, 0.99]} & $0.750$ & $0.380$ \\
        \quad Mutual roles & Binary & \textbf{0.90} \tiny{[0.84, 0.96]} & 0.89 \tiny{[0.81, 0.95]} & 0.89 \tiny{[0.81, 0.95]} & $0.872$ & $0.872$ \\
        \quad Respect privacy & Binary & \textbf{1.00} \tiny{[1.00, 1.00]} & 0.95 \tiny{[0.89, 0.99]} & 0.96 \tiny{[0.92, 0.99]} & $0.750$ & $0.712$ \\
        \quad Partnership building & Binary & 0.66 \tiny{[0.55, 0.76]} & 0.55 \tiny{[0.44, 0.65]} & \textbf{0.74} \tiny{[0.65, 0.83]} & $0.750$ & $0.299$ \\
        \quad Caring \& commitment & Binary & \textbf{0.97} \tiny{[0.94, 1.00]} & 0.86 \tiny{[0.80, 0.93]} & 0.89 \tiny{[0.83, 0.94]} & $0.690$ & $0.179$ \\
        \quad Acknowledge mistakes & Binary & \textbf{0.98} \tiny{[0.94, 1.00]} & 0.90 \tiny{[0.81, 0.98]} & 0.91 \tiny{[0.84, 0.98]} & $0.750$ & $0.860$ \\
        \quad Greet appropriately & Binary & \textbf{1.00} \tiny{[1.00, 1.00]} & 0.97 \tiny{[0.93, 1.00]} & 0.97 \tiny{[0.93, 1.00]} & $0.750$ & $0.872$ \\
        \quad Appropriate language & Binary & \textbf{0.99} \tiny{[0.97, 1.00]} & 0.95 \tiny{[0.90, 0.98]} & 0.95 \tiny{[0.90, 0.99]} & $0.750$ & $0.713$ \\
        \quad Encourage participation & Binary & \textbf{0.95} \tiny{[0.91, 0.99]} & 0.92 \tiny{[0.86, 0.97]} & 0.91 \tiny{[0.85, 0.96]} & $0.750$ & $0.828$ \\
        \quad Interest in person & Binary & 0.67 \tiny{[0.57, 0.76]} & 0.57 \tiny{[0.46, 0.66]} & \textbf{0.77} \tiny{[0.67, 0.84]} & $0.690$ & $0.153$ \\
        \addlinespace[0.25em]
        \multicolumn{7}{@{}l}{\textbf{\textsf{Clinical \& Communication Competencies (PACES)}}} \\
        \quad Address concerns & 5-point Likert & \textbf{0.87} \tiny{[0.83, 0.91]} & 0.86 \tiny{[0.82, 0.90]} & 0.81 \tiny{[0.75, 0.85]} & $0.472$ & $0.482$ \\
        \quad Confirm understanding & 5-point Likert & \textbf{0.68} \tiny{[0.61, 0.74]} & 0.66 \tiny{[0.59, 0.72]} & 0.65 \tiny{[0.57, 0.72]} & $0.750$ & $0.482$ \\
        \quad Empathy & 5-point Likert & \textbf{0.76} \tiny{[0.71, 0.80]} & 0.70 \tiny{[0.65, 0.74]} & 0.69 \tiny{[0.63, 0.74]} & $0.472$ & $0.128$ \\
        \quad Maintain welfare & 5-point Likert & \textbf{0.85} \tiny{[0.81, 0.89]} & 0.84 \tiny{[0.79, 0.88]} & 0.83 \tiny{[0.79, 0.88]} & $0.830$ & $0.828$ \\
        \addlinespace[0.25em]
        \multicolumn{7}{@{}l}{\textbf{\textsf{Interface \& Modality Preference}}} \\
        \quad Effective comm. & 5-point Likert & 0.89 \tiny{[0.85, 0.92]} & 0.79 \tiny{[0.75, 0.82]} & \textbf{0.90} \tiny{[0.86, 0.93]} & $0.750$ & $4.98 \times 10^{-5}$ \\
        \quad Felt understood & 5-point Likert & \textbf{0.90} \tiny{[0.86, 0.93]} & 0.81 \tiny{[0.77, 0.85]} & 0.88 \tiny{[0.83, 0.92]} & $0.830$ & $2.62 \times 10^{-4}$ \\
        \quad Easy \& convenient & 5-point Likert & 0.88 \tiny{[0.84, 0.92]} & 0.71 \tiny{[0.66, 0.77]} & \textbf{0.90} \tiny{[0.86, 0.93]} & $0.750$ & $2.98 \times 10^{-6}$ \\
        \bottomrule
    \end{tabular*}
\end{table}
\clearpage
\section{Full Patient Actor Preferences}
\label{appendix:full_patient_actor_ratings}

\begin{figure}[h!]
    \centering
    \includegraphics[width=\textwidth]{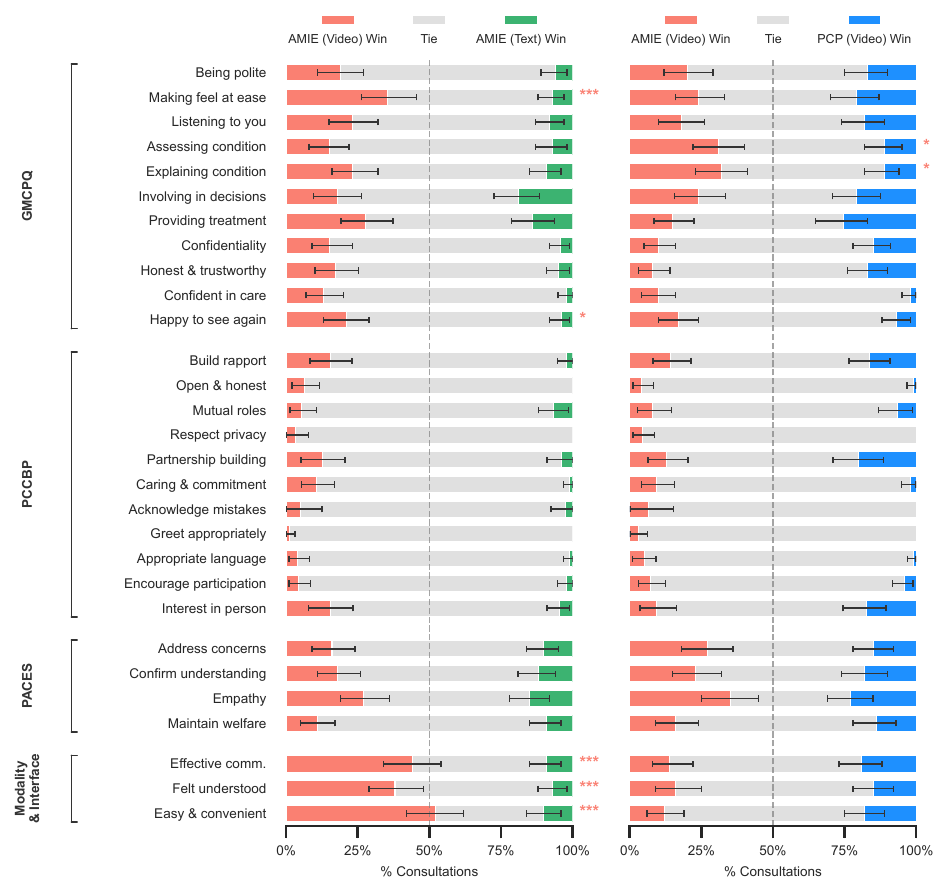}
    \vspace{0.1cm}
    \caption{\textbf{Patient Actor Preferences.} \textbf{Left:} Preferences (per scenario) for each criterion between AMIE (Video) in red and AMIE (Text) in green. \textbf{Right:} Preferences (per scenario) for each criterion between AMIE (Video) in red and PCP in blue. Error bars correspond to 95\% confidence intervals from bootstrapping (n=1000). For statistical comparison, a two-sided Wilcoxon signed-rank test with Benjamini--Hochberg false discovery rate (FDR) correction was used to compare AMIE (Video) to the other arms (adjusted $p$-values: $*p < 0.05$, $**p < 0.01$, $***p < 0.001$).}
    \label{figure:patient_actor_rating_preferences}
\end{figure}
\clearpage
\section{Full Expert Evaluation Preferences}
\label{appendix:full_expert_eval_ratings}

\begin{figure}[h!]
    \centering
    \includegraphics[width=0.92\textwidth]{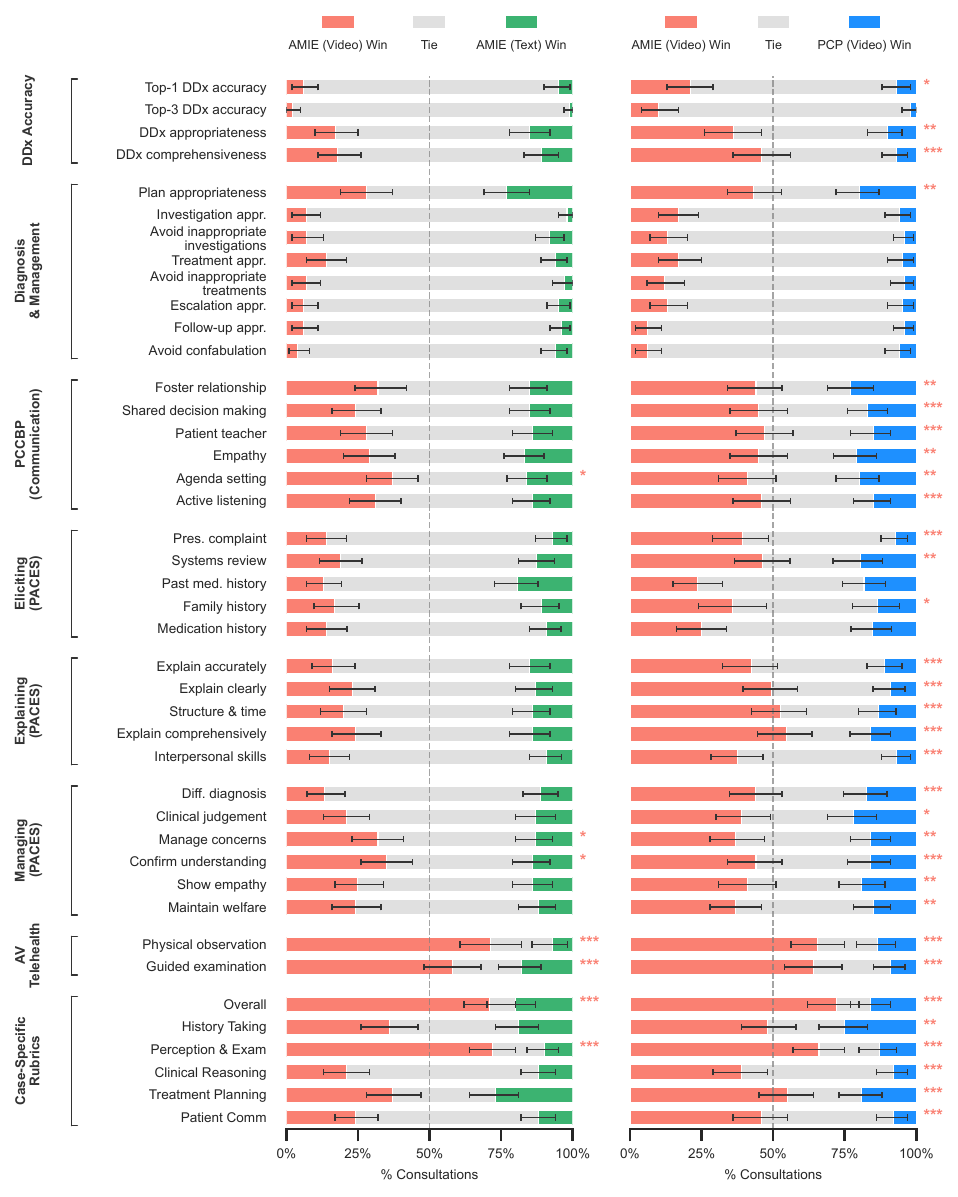}
    \caption{\textbf{Expert Preferences.} \textbf{Left:} Preferences (per scenario) for each criterion between AMIE (Video) in red and AMIE (Text) in green. \textbf{Right:} Preferences (per scenario) for each criterion between AMIE Video in red and PCP in blue. Error bars correspond to 95\% confidence intervals from bootstrapping (n=1000). For statistical comparison, a two-sided Wilcoxon signed-rank test with Benjamini--Hochberg false discovery rate (FDR) correction was used to compare AMIE (Video) to the other arms (adjusted $p$-values: $*p < 0.05$, $**p < 0.01$, $***p < 0.001$).}
    \label{figure:expert_rating_preferences}
\end{figure}

\clearpage
\section{Patient Actor Interviews}
\label{appendix:patient_actor_interviews}

To complement the quantitative findings, we conducted a  qualitative evaluation with 7 patient actors to understand how communication modalities (text, audio, video) impact the perceived quality of care, the effectiveness of health plan communication, and the overall dynamics of the clinical encounter. The interviews lasted 25 minutes and were recorded and transcribed. We used reflexive thematic analysis to inductively generate themes based on the transcripts and researcher notes. The analysis yielded three distinct themes:
\begin{itemize}
    \item \textbf{Modality suitability depends on context:} Clinical needs and environmental factors dictate the ideal interface, suggesting there is no “one-size-fits all” modality.
    \item \textbf{Though empathetic, AI lacks conversational cadence:} AI provides consistent empathy, breaks down complex medical jargon for patients, and is thorough in its evaluation of the patient, but it lacks natural conversational capabilities.
    \item \textbf{The need for human connection is dictated by symptom urgency and severity:} The severity of a patient’s condition shifts their trust requirements in AI. 
\end{itemize}
Exploring these themes revealed that the video modality is preferred for certain domains such as musculoskeletal (MSK) exams because it allows participants to visually demonstrate range of motion and point out locations where they are feeling pain. In contrast, the text modality affords an anonymity that allows participants to feel less self-conscious and ensures privacy when discussing sensitive neurological/psychiatric topics. The need for text is also dictated by environmental factors; when in public settings or when voice communication is difficult, text becomes advantageous for maintaining privacy. For the dynamics of the clinical interaction itself, human doctors remain the “gold standard” for social connection, but participants agreed they still felt heard and were unhurried by the AI. Assessing the different modalities, participants agreed that for minor issues, the AI would serve as an effective 24/7 triage tool, but for more serious or life-threatening conditions, they wanted human connection requiring real-time audio/video interactions.

\subsection{Overview}

The goal of the user experience research (UXR) study was to evaluate the participant experience across different modalities (text, audio, video) within the audio-video OSCE framework. Through remote, moderated interviews, UXR answered the following research questions:

\begin{itemize}
    \item What are the unique benefits and drawbacks of text, audio-only, and video interactions for both diagnosis and care management?
    \item To what extent does the presence or absence of visual feedback (synchronous video v. “blank screen” audio-only) influence participant perceptions of clinician/AI empathy during care encounters?
    \item Which communication modality (text, audio, video) are naturally prioritized and why?
\end{itemize}

\subsection{Methodology}

\subsubsection{Methods}

UXR conducted remote, moderated semi-structured interviews to collect insights from patient actors about their perceived quality of care and overall dynamics of the clinical encounters they engaged in. 

\subsubsection{Data Collection Process}

Data was collected with 7 patient actors who were part of the larger AV-AMIE OSCE study. Data collection was completed remotely over Google Meet by the manager of the OSCE lab. During the sessions, participants were asked questions to better understand their thought processes regarding the value and experience of each modality. Participants were prompted to reflect on how their ability to convey health information varied based on whether they were using text, audio or video. They also discussed how the absence of video, a “blank screen,” influenced their perception of the provider’s empathy during the interaction. 

\subsubsection{Data Analysis}

Interview notes were collected using the interview transcripts, and the data was analyzed using reflexive thematic analysis. The open-ended interview responses were coded first with an initial codebook designed to capture core concepts. The codebook was adjusted and refined, and once notes for all 7 participants were coded, the codes were synthesized and grouped to identify overarching themes.

\subsection{Key Findings}

Thematic analysis of the interview sessions generated three themes about how patients navigate, evaluate, and prioritize multimodal AI clinical interactions based on clinical context and how each modality supports them

\subsubsection{Modality Suitability Depends on Context}

Clinical needs and environmental factors dictate the ideal interface suggesting there is not a single modality that is preferred for any given condition. When participants seek a physical evaluation, video is the preferred modality as it allows participants to point out where pain occurs or demonstrate range of motion. For example, for musculoskeletal (MSK) concerns, participants shared that a video modality would allow pointing to the site of pain or help demonstrate range of motion or perform other movements.

Audio comes second to video --- it allows for the visit to feel personable and comforting, allowing participants to build rapport with the provider when text cannot. Still, it lacks the visual connection which provides the conversation the holistic picture needed for certain conditions, such as musculoskeletal conditions. 

Finally, when seeking anonymity for sensitive topics like mental health, participants agree that text offers the additional privacy they need. P5 noted that for sensitive topics, such as reproductive health, using the text modality would afford a level of privacy not available in modalities that require speaking. In addition to the context of the condition, environmental constraints also dictate preferred modality. Text is preferred for receiving medical advice in public spaces when voice communication is difficult, e.g. seeking out care on a busy train ride.

\subsubsection{Though Empathetic, AI Lacks Conversational Cadence}

Participants conveyed that while the AI breaks down complex medical jargon for patients, provides empathy, and is thorough in its evaluation of symptoms, especially for non-text modalities, it still lacks natural conversation capabilities. The rapport and empathy built with human providers varies based on their schedules. One participant observed that human providers’ empathy mimicked “peaks and valleys” depending on the day, but the AI is consistent in its empathy. The AI also did not rush the participants --- there is no pressure to condense information in order to hurry the conversation along because there is another patient in the waiting room.

However, unlike the human connection with a human provider where small talk was commonplace, participants reported that conversations with the AI felt awkward at times. Participants shared that the natural cadence of a conversation was lacking, making it obvious they were speaking to AI and not a human. The AI also presented participants with awkward silences causing some participants to feel uncomfortable or confused as to whether or not they were still connected to the AI agent and prompting suggestions of how to demonstrate that the system is still processing.

\subsubsection{The Need for Human Connection Is Dictated by Symptom Urgency and Severity}

Modalities for care vary based on the urgency and severity of the symptoms a participant faces. For acute, minor conditions, participants trust text modalities to help triage symptoms and determine whether or not immediate, in-person care is needed. For straightforward, non-urgent concerns, like a sinus infection, the AI is efficient and thorough.

However, when dealing with more serious, life-threatening and urgent health concerns, participants prefer video, or audio, visits to seek out the human voice and feel a sense of trust and security. For example, one participant shared a preference for video when they were faced with elevated blood pressure. In these emergent situations, participants felt ease in establishing rapport with and hearing from a human provider.

\subsection{Conclusion}

These findings provide insights into the information needs and modality preferences of patient actor participants engaging with multimodal AI clinical tools. Overall, the data reveals that there is no universal, one-size-fits-all modality. Instead, patient preferences are situational and dependent on the symptoms they are exhibiting, the severity of the symptoms, and their environmental constraints. The AI in this study was recognized to be effective for the initial screening of symptoms, and participants reasoned they would use it for triaging their care or seeking a tool for quick advice when immediate care is unavailable.

Future iterations of multimodal AI tools for health would benefit from addressing conversational limitations. Here, while patient actor participants appreciated the system’s ability to break down medical jargon into accessible language and to provide consistent empathy, the lack of natural conversational cadence made the interactions feel awkward and unnatural. While clinical AI tools were seen to have value as a triage tool, the perspective was that human providers, via video or in-person, continue to be the “gold standard” for care.

\clearpage
\begingroup
\tiny
\begin{longtable}{
  p{0.12\textwidth} 
  p{0.15\textwidth} 
  p{0.12\textwidth} 
  p{0.15\textwidth} 
  p{0.18\textwidth} 
  p{0.23\textwidth}
}

\caption{Qualitative interview themes from patient participants.}
\label{tab:qual-themes} \\
\toprule
\textbf{Theme} & \textbf{Definition} & \textbf{Code Name} & \textbf{Definition} & \textbf{Inclusion Criteria} & \textbf{Quotes} \\
\midrule
\endfirsthead

\multicolumn{6}{c}%
{{\tablename\ \thetable{} -- continued from previous page}} \\
\toprule
\textbf{Theme} & \textbf{Definition} & \textbf{Code Name} & \textbf{Definition} & \textbf{Inclusion Criteria} & \textbf{Quotes} \\
\midrule
\endhead

\midrule
\multicolumn{6}{r}{{Continued on next page}} \\
\endfoot

\bottomrule
\endlastfoot

Modality suitability depends on context & Clinical symptoms, immediate environmental/privacy factors, and nature of the condition dictate communication modality (text, audio, video)
    & Modality -- Text & Preferences, benefits, or drawbacks associated with text-only communication & Typing effort, response speed, public privacy, clear articulation, lack of emotion, control, redundant replies, or cold/impersonal text feelings & ``\ldots text AI feels like a thorough Google search, but it is highly useful for privacy when in public settings like a train.'' --P2 ``thought with the text the AI was superior in my eyes\ldots the AI was very quick to get back to you\ldots texting human doctors was a long process with frustrating wait times between messages.'' --P4 \\
\addlinespace

  &  & Modality -- Video & Preferences, benefits, or drawbacks associated with video communication & Physical movements, pain locations, efficiency, visual confirmation, self-consciousness, seeing one's reflection, missing provider's face hindering rapport, or preferring a blank screen over an artificial AI persona & ``I thought it was easier to use the video um option with the AI doctor just because it felt like it was easier to communicate, express myself.'' --P5 ``I could show myself, maybe go through physical exam stuff that I wasn't able to do in a text or audio format\ldots'' --P7 \\
\addlinespace

  &  & Modality -- Audio & Preferences, benefits, or drawbacks associated with audio communication & Hearing a voice, establishing a connection for serious conditions, tone, comfort/ease from the voice, easy verbal description, or audio improving real-time interaction & ``the AI voice, I felt like it improved throughout the course of doing um like the sessions\ldots I feel like hearing a voice, having the tone of a voice is something that I actually really like.'' --P6 ``audio\ldots was a little easier to express myself and it just felt more like a real interaction being able to hear another voice on the other end.'' --P7 \\
\addlinespace

  &  & MSK/Physical Exams & Issues related to musculoskeletal conditions and the need for physical demonstration & Pointing to pain, demonstrating range of motion (rotation, gait), isolating muscles, skepticism of AI interpreting movement, observation vs.\ human nuance, physicians rarely asking for exams, or uncertainty of AI capture & ``The text one they just asked me to describe what happened when I did it\ldots but when I could show them\ldots I felt more confident because they're actually seeing it\ldots'' --P3 ``I think a video for these because you can do specific movements versus like describing like I'm supinating my arm and it hurts. You can just show them and then they can interpret that.'' --P6 \\
\addlinespace

  &  & Mental Health \& Privacy & Issues related to psychological conditions and the desire for privacy or reduced intimidation & Anxiety, depression, text preference to avoid self-consciousness, public privacy (e.g., trains), text feeling safer / comfortable, or anonymity easing sensitive mental / reproductive health discussions & ``With maybe more sensitive issues, like if I was having reproductive symptoms that I didn't really know how to ask about, I would probably do text AI so that people wouldn't hear me.'' --P5 ``Preferred text for psych-related scenarios because the sense of anonymity made it easier to discuss sensitive topics.'' --P7 \\
\midrule

Though empathetic, AI lacks conversational cadence & Clinical AI provides a thorough, standardized assessment and is empathetic, but is conversationally stiff and robotic, lacking natural conversational qualities
    & System Feedback \& Silences & User reactions to system processing time and the need for visual or auditory cues & Buffering, loading, pauses, awkward silences, AI processing indicators (wheel, light), thinking time from silences, connection/crash worries, or ambiguity of input receipt & ``Silences caused worry that the connection had failed, especially with the AI.'' --P5 ``Silences created ambiguity about whether the system received or understood the input.'' --P7 \\
\addlinespace

  &  & Naturalness of Interactions & The degree to which the interaction feels like a natural human conversation & Call-and-response cadence, robotic feel, conversational flow, human interaction comparisons, or human elements like jokes, small talk, and past context & ``PCP is the gold standard due to human connection. Video was easier for communication and expressing oneself.'' --P5 ``Audio was a step up for natural interaction\ldots not seeing the provider's face made it hard to build a connection.'' --P7 \\
\addlinespace

  &  & Empathy \& Connection & The perceived emotional support, warmth, and feeling of being understood by the provider (AI or human) & Feeling heard, active listening, symptom summaries, warm feeling, tone, lack of genuine emotion, AI empathy consistency vs.\ human variability, or feeling at ease & ``Primary care physician is the gold standard because human connection is tough to replicate\ldots natural cadence, jokes, remembering past context,'' --P2 ``I thought AI was like more consistent with providing the right amount of empathy with each different case, whereas human physicians were inconsistent, sometimes feeling cold.'' --P5 \\
\addlinespace

  &  & Thoroughness \& Diligence & The perceived completeness and detail of the medical examination and questioning & AI follow-up, symptom summaries, not feeling rushed, due diligence, observing incidental symptoms (coughs/sneezes), or AI's thoroughness vs.\ rigid human doctor scripts & ``I did not feel rushed. Despite being shorter, the AI was thorough and always asked if there was anything else to add before concluding.'' --P3 ``I thought AI was actually better at like wanting to do a physical exam over video or text. Like I thought they were more proactive in doing that.'' --P5 \\
\addlinespace

  &  & Medical Communication \& Guidance & How medical information, instructions, and terminology are conveyed to the patient & Simplifying jargon, actionable safety steps (unlocking doors, meds), guiding movements, clear explanations for lower literacy, or AI recommending urgent care & ``The AI did a good job explaining some of this terminology to me\ldots broke down complex jargon\ldots'' --P1 ``Human PCPs provide anecdotal treatment advice\ldots AI appeared more cautious, frequently recommending urgent care.'' --P2 \\
\addlinespace

  &  & Visit Duration & Visit length and perceived time pressure during the encounter & Visit length, feeling rushed by human doctors (quotas/waiting rooms), taking time with AI, AI repeating/dragging, or text vs.\ video efficiency & ``Video is definitely more efficient (5--10 mins) compared to texting (15--20 mins). Compared to a human PCP, AI video was similar or slightly shorter.'' --P2 ``AI text and video sessions typically lasted 10--15 minutes, which felt comfortable\ldots Felt more rushed during interactions with human primary care physicians.'' --P5 \\
\midrule

The need for human connection is dictated by symptom urgency and severity & The severity and urgency of clinical symptoms dictate when human connection is required for certain health concerns (e.g.\ life threatening requiring human connection vs.\ triaging by AI)
    & Trust \& Safety & User confidence in the diagnosis and the perceived caution of the provider & AI caution, urgent care recommendations, wanting human confirmation, AI as a triage gateway (after-hours/night, ER decisions), or saving on in-person care costs & ``\ldots readily use AI as a triage gateway for after-hours needs, quick advice, or to decide whether to go to the ER, especially for serious things in the middle of the night.'' --P3 ``\ldots would prioritize using the AI video doctor first as a triage step to save on expensive in-person care.'' --P5 \\
\addlinespace

  &  & Emergency \& Serious Conditions & Preferences for handling urgent, severe, or life-threatening health concerns & Strokes, heart attacks, high blood pressure, needing visual/audio connection, tone urgency, AI safety instructions (unlocking doors, meds), or ER/urgent care decisions & ``high blood pressure scare\ldots preferred video for speed to get point across quickly in a potential emergency\ldots typing out detailed information was more time-consuming and stressful.'' --P1 ``For a serious condition like a stroke, I wanted to hear a voice and establish a connection. Text felt inadequate\ldots I really wanted to talk to somebody.'' --P3 \\

\end{longtable}
\endgroup

\newpage
\setlength\bibitemsep{3pt}
\printbibliography
\clearpage
\end{refsection}

\end{document}